\documentclass[manuscript, nonacm]{acmart}
\usepackage{multirow}
\usepackage{pifont}
\usepackage{makecell}
\usepackage{algorithm}
\usepackage{algorithmic}
\usepackage{amsmath}
\usepackage{subcaption}
\usepackage{mathtools}
\usepackage[utf8]{inputenc}
\DeclareMathOperator*{\argmax}{arg\,max}
\DeclareMathOperator*{\argmin}{arg\,min}

\AtBeginDocument{%
  }

\begin{document}

\title{SGPO: Self-Generated Preference Optimization based on Self-Improver}

\author{Hyeonji Lee}
\affiliation{%
  \institution{Korea University}
  \city{Seoul}
  \country{Republic of Korea}}
\email{0105lhl@korea.ac.kr}
\orcid{0009-0000-7070-209X}

\author{Daejin Jo}
\affiliation{%
  \institution{Korea University}
  \city{Seoul}
  \country{Republic of Korea}}
\email{twidddj@gmail.com}
\orcid{0009-0008-3778-7798}

\author{Seohwan Yun}
\affiliation{%
  \institution{Korea University}
  \city{Seoul}
  \country{Republic of Korea}}
\email{jitoo6342@korea.ac.kr}
\orcid{0009-0008-1831-1559}

\author{Sungwoong Kim}
\affiliation{%
  \institution{Korea University}
  \city{Seoul}
  \country{Republic of Korea}}
\email{swkim01@korea.ac.kr}
\orcid{0000-0002-2676-9454}
\authornote{Corresponding author. Email: swkim01@korea.ac.kr}

\renewcommand{\shortauthors}{H. Lee et al.}

\begin{abstract}
Large language models (LLMs), despite their extensive pretraining on diverse datasets, require effective alignment to human preferences for practical and reliable deployment. 
Conventional alignment methods typically employ off-policy learning and depend on human-annotated datasets, which limits their broad applicability and introduces distribution shift issues during training.
To address these challenges, we propose \textbf{S}elf-\textbf{G}enerated \textbf{P}reference \textbf{O}ptimization based on Self-Improver \textbf{(SGPO)}, an innovative alignment framework that leverages an on-policy self-improving mechanism. Specifically, the improver refines responses from a policy model to self-generate preference data for direct preference optimization (DPO) of the policy model. Here, the improver and policy are unified into a single model, and in order to generate higher-quality preference data, this self-improver learns to make incremental yet discernible improvements to the current responses by referencing supervised fine-tuning outputs. Experimental results on AlpacaEval 2.0 and Arena-Hard show that the proposed SGPO significantly improves performance over DPO and baseline self-improving methods without using external preference data.
\end{abstract}



\maketitle

\section{Introduction}
Large language models (LLMs) acquire broad world knowledge by pretraining on massive corpora through an objective of auto-regressive next-token prediction. 
While such pretraining enables impressive generalization capabilities, aligning LLMs with human preferences, referred to as human alignment, is crucial for safe and useful deployment in real-world applications \cite{achiam2023gpt, team2023gemini, grattafiori2024llama}.
To this end, a variety of alignment algorithms, most notably in forms of Reinforcement Learning from Human Feedback (RLHF) \cite{ouyang2022training} and Direct Preference Optimization (DPO) \cite{rafailov2023direct}, have been recently developed based on a preference modeling framework such as the Bradley-Terry model \cite{bradley1952rank}.

Especially, preference optimization approaches based on the DPO method have been widely used in recent alignment research, since they directly optimize the language model without requiring the training of a separate reward model, simplifying the optimization process.
These DPO-based  methods have shown remarkable success in aligning LLMs by fine-tuning them to prefer human-endorsed responses over less suitable responses.
However, they critically rely on high-quality human preference datasets, the construction of which is often prohibitively expensive and time-consuming \cite{christiano2017deep, ouyang2022training}.
In practice, one may resort to publicly available preference datasets to reduce costs, but the distribution mismatch between the behavior policy that was used to collect these preference datasets and the target policy we aim to update can significantly degrade the effectiveness of alignment, restricting the model's improvement potential \cite{li2023policy, lin2024limited}.

As a result, there is growing interest in methods that reduce or eliminate the dependence on manually curated preference datasets by automating the alignment process \cite{chen2024self, lee2024aligning}.
Among them, SPIN \cite{chen2024self} proposes a self-improvement mechanism in which a model refines itself through self-play: it generates responses from earlier checkpoints of an LLM and learns to distinguish and prefer stronger responses over time, utilizing supervised fine-tuning data (SFT). 
However, since SPIN directly takes off-policy and static SFT responses as target preferred responses, there remains a distributional gap between the self-generated responses and the supervised references, which can limit the effectiveness of self-improvement.
On the other hand, for self-improving preference optimization, SynPO \cite{dong2024self} introduces a fully synthetic alignment framework where a prompt generator and a separate response improver—distinct from the main LLM policy—work in tandem to create diverse prompts and refined responses. Here, the improver is trained using demonstrations collected from an external LLM (e.g., GPT-4 Turbo \cite{openai_gpt4turbo}), conditioned on self-generated responses as inputs. 
Although, compared to SPIN, SynPO enables a more on-policy-like learning process in that both the preferred (chosen) and dispreferred (rejected) responses used for optimization are directly derived from the current policy's outputs, the improver is trained separately from the policy. Therefore, as one of them gets updated, the improver would diverge from the policy and lack access to the internal state or behavior of the current policy, making it difficult to fully realize a true on-policy modeling.

To address these limitations, we propose the use of an on-policy self-improver—a mechanism designed to align the refinement process more closely with the current policy. 
Specifically,
\textbf{1)} to ensure on-policy behavior, it is beneficial to unify the improver and policy into a single shared model architecture.
By sharing parameters, the improver inherently reflects the current policy’s internal representations and decision patterns, making the refinement process more policy-aware.
From the perspective of preference optimization, such a unified model can serve both roles by adjusting the input prompts, eliminating the need for a separate improver.
\textbf{2)} Moreover, to facilitate more stable and effective learning, it is important to generate improved responses that are achievable from the current policy. As suggested by prior work~\cite{wu2024beta}, overly aggressive refinements can introduce large preference gaps that hinder effective optimization; in contrast, slight yet consistent improvements allow the policy to learn more efficiently.
\textbf{3)} Finally, to provide a clear learning signal, the improver should be trained to refine responses in the direction of high-quality outputs. This can be achieved by referencing SFT responses, guiding refinements by the improver toward desired responses.

Based on these considerations, we introduce \textbf{S}elf-\textbf{G}enerated \textbf{P}reference \textbf{O}ptimization based on Self-Improver \textbf{(SGPO)}, an effective LLM alignment framework designed to achieve the three objectives mentioned above. In particular, we employ a shared model integrating both the policy model and the improver model to self-generate preference pairs that are more on-policy for direct preference optimization. Moreover, the self-improver refines the current policy's responses conditioned on SFT responses for clear improvements while maintaining attainable improvements between chosen and rejected samples for effective training.
To demonstrate the effectiveness of SGPO trained with the proposed self-generated pairwise preference data, we evaluate its performance on well-known benchmarks with various LLMs. The results show that SGPO consistently and significantly outperforms baseline preference optimization methods including DPO and previous self-improving approaches. Especially, it improves the performance over DPO by up to 16.02\% on the AlpacaEval 2.0 \cite{alpaca_eval} in terms of length control win rate and by up to 17.3\% on Arena-Hard \cite{arenahard2024, li2024crowdsourced} in terms of win rate when applied to Qwen2.5-Base (7B) \cite{team2024qwen2}. For Llama3-Base (8B) \cite{grattafiori2024llama}, SGPO achieves better performance up to 6.28\% and 13.9\% on the same benchmarks, respectively, even without using external preference data. To summarize, our main contributions are as follows:
\begin{enumerate}
    \item We propose SGPO, a self-improving preference optimization framework that employs an on-policy self-improver to generate preference data for direct preference optimization.
    
    \item We integrate the policy and improver into a unified model, efficiently generating more online preference data through simple prompt modifications.

    \item Our self-improver is trained to generate high-quality preference data by constructing gradual improvements to current responses based on SFT responses.
    
    \item Experimental results demonstrate substantial performance improvements by SGPO over existing baselines with various LLMs and benchmarks, without using external preference supervision.
\end{enumerate}

\begin{figure*}[t]
  \centering
  \includegraphics[width=1\linewidth]{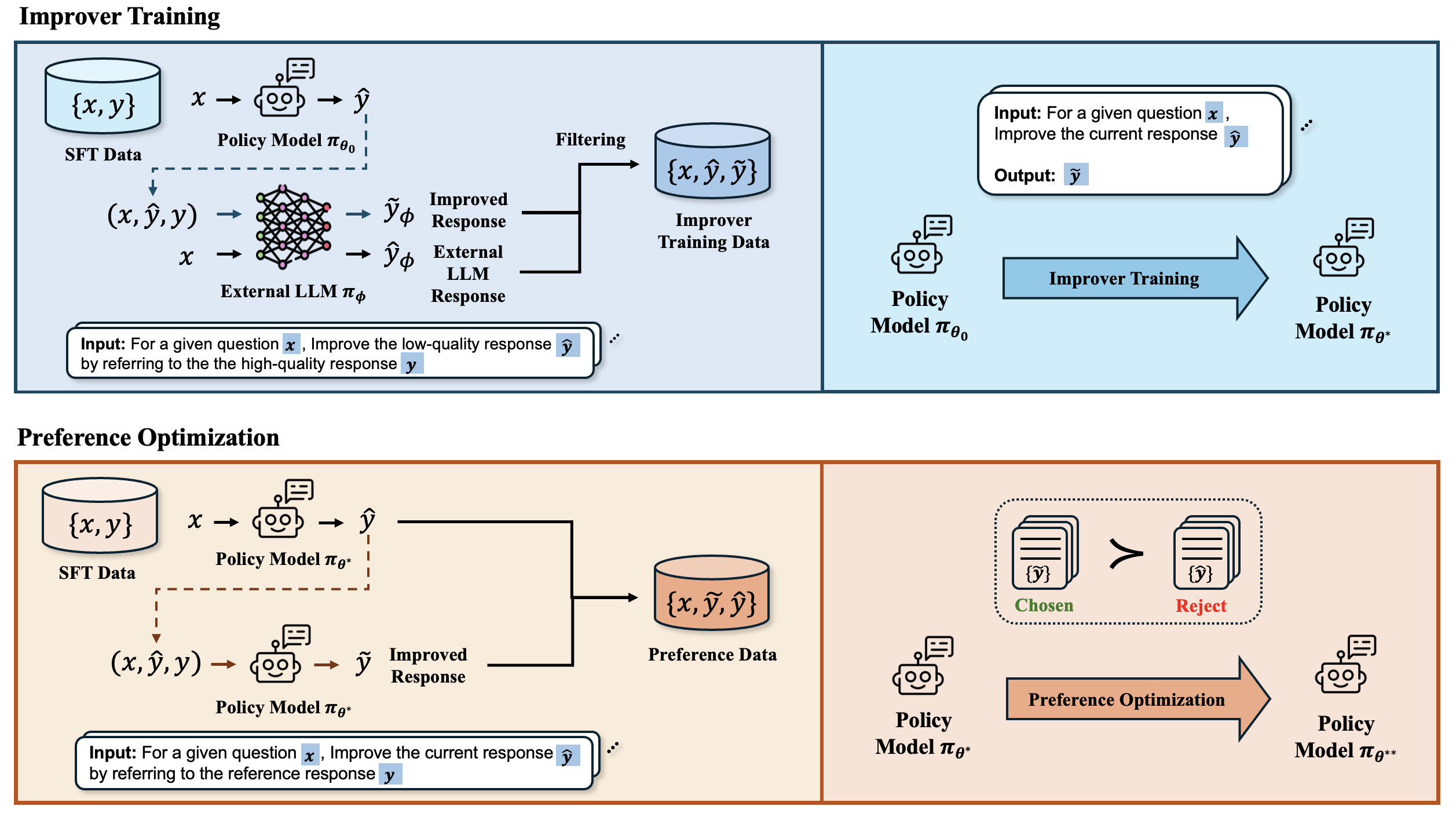}
  \caption {Overview of the Self-Generated Preference Optimization (SGPO) framework. The SGPO framework consists of two steps: improver training and preference optimization. In Step 1, the initial policy model $\pi_{\theta_{0}}$ is used to construct training data for the improver, enabling it to refine responses and update the model to $\pi_{\theta^*}$. In Step 2, preference optimization is performed using online preference data, incorporating both the improved responses and policy responses generated by $\pi_{\theta^*}$, resulting in the final model update.}
  \label{fig1}
  \Description{Figure 1. Fully described in the text.}
\end{figure*}

\section{Related Work}
\subsection{Direct Alignment Algorithms.} Direct Preference Optimization (DPO) method \cite{rafailov2023direct}, widely used for human alignment, is a representative offline preference optimization approach. Instead of training an explicit reward model, DPO directly updates the policy using preference data. Building on this foundation, various direct alignment methods such as SimPO \cite{meng2024simpo} and ORPO \cite{hong2024orpo} have been proposed. SimPO simplifies preference optimization by using average log probabilities as implicit rewards, removing the need for a reference model, and ORPO is a simple odds ratio-based method that achieves preference alignment by applying small penalties during the supervised fine-tuning. In addition, there exist methodologies such as KTO \cite{ethayarajh2024kto}, which proposes directly maximizing the utility of generations instead of the log-likelihood of preferences, eliminating the need for preference pair datasets, and $\beta$-DPO \cite{wu2024beta}, which analyzes the influence of preference data quality and the parameter $\beta$ for DPO training and introduces a method that dynamically adjusts $\beta$ at the batch level based on data quality. However, these direct alignment methods require a large amount of high-quality preference data, which is costly and time-consuming, and often induce off-policy learning problems.
\subsection{Self-Improving Alignment.} Recent studies have increasingly focused on on-policy approaches in which LLMs are self-improved from their own responses. For example, SPIN \cite{chen2024self} adopts a self-play mechanism that iteratively generates training data using the LLM itself, enhancing performance in conjunction with an SFT dataset, while APO \cite{cheng2023adversarial} proposes an adversarial preference optimization framework where the LLM and the reward model are alternately updated in a min-max game setting during the alignment. Self-rewarding approaches \cite{wu2024meta} exploit the LLM-as-a-Judge paradigm, guiding the model to provide its own rewards during training. SynPO \cite{dong2024self} introduces a self-boosting mechanism to generate synthetic preference data. It employs a response refiner to refine current policy responses for direct preference optimization. However, these self-improving alignment methods have inherent limitations from the perspectives of their restricted online-ness of the improved data or an uncontrolled improvement gap between the chosen and rejected samples.

\begin{figure}[h]
    \centering
    {\includegraphics[width=0.9\linewidth]{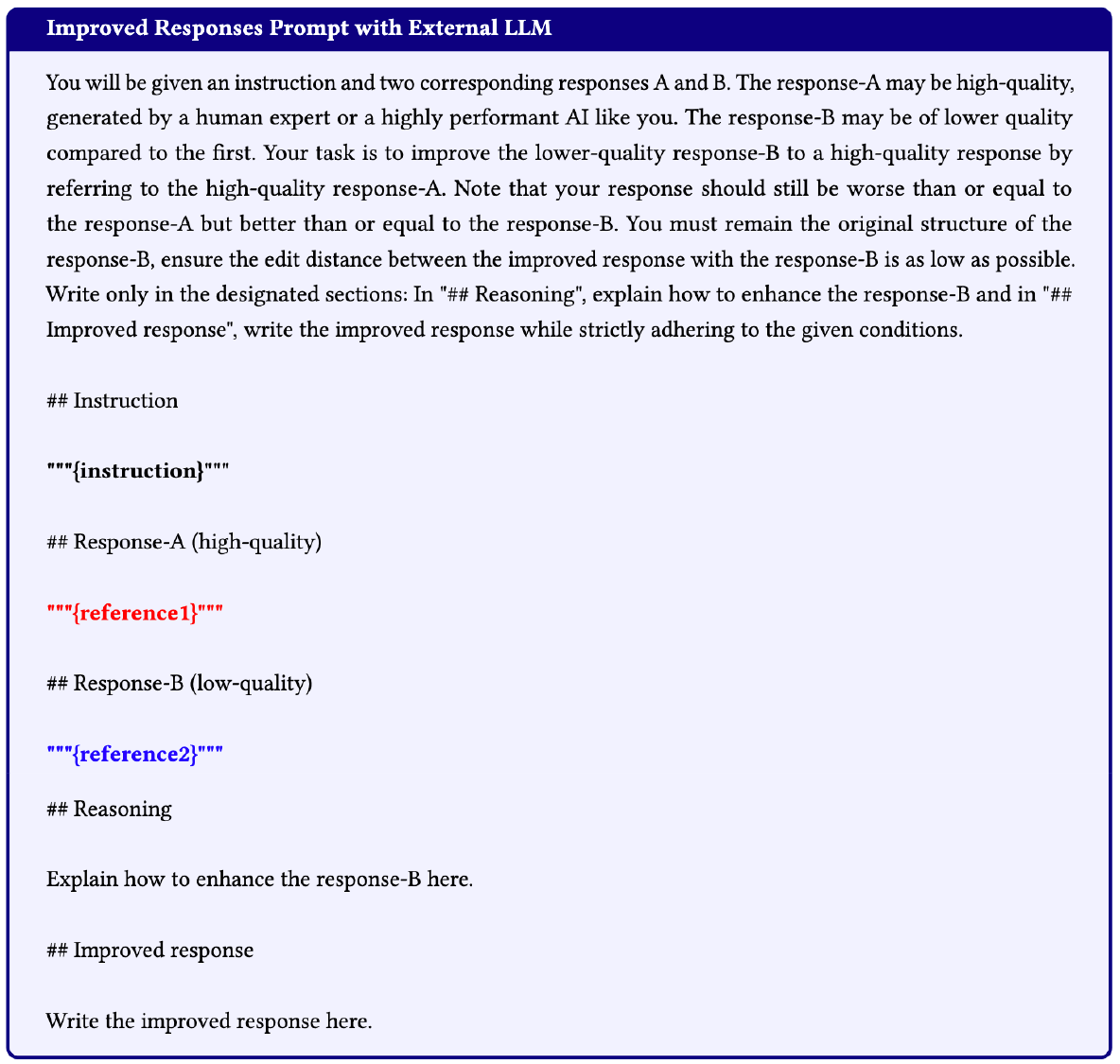}}
    \caption {Prompt for an external LLM to obtain target improved responses.}
    \label{fig5_prompt2}
    \Description{Figure 2. Fully described in the text.}
\end{figure}

\section{Method} 
\label{sec:Method}
The proposed SGPO is an on-policy alignment framework designed to generate high-quality preference data without relying on human-annotated comparisons. The overall framework of SGPO is depicted in Figure~\ref{fig1}.

\subsection{On-Policy Self-Improver}
Given SFT data for the policy update $S_P= \{(\mathbf{x}_{j},\mathbf{y}_{j})\}_{j=1}^{m}$, the proposed self-improver refines a response ${\hat{\bf y}}_{\theta,j}$ of the current policy $\pi_{\theta}$ for an input ${\bf x}_j$ and produces an improved response $\mathbf{\tilde y}_{\theta,j}$ using the same $\pi_{\theta}$ with a refinement prompt $r$ such that
\begin{equation}\label{equation 1}
  \begin{aligned}
    \hat{\mathbf{y}}_{\theta,j} &\sim \pi_{\theta}\bigl(\cdot \mid \mathbf{x}_j\bigr),\\
    \tilde{\mathbf{y}}_{\theta,j} &\sim \pi_{\theta}\bigl(\cdot \mid r(\mathbf{x}_j, \mathbf{y}_j, \hat{\mathbf{y}}_{\theta,j})\bigr).
  \end{aligned}
\end{equation}
Here, the improved response $\mathbf{\tilde y}_{\theta,j}$ is generated with reference to the SFT response $\mathbf{y}_j$ to ensure a clear improvement.

In order the policy to act as the improver as well, we perform the improver training by constructing a dedicated training data ${\mathcal D}_R$. Specifically, given a separate small SFT dataset for improver training $S_R=\{(\mathbf{x}_{i}, \mathbf{y}_{i})\}_{i=1}^{n}$ ($n << m$), for each ${\bf x}_i$, we first generate a response $\mathbf{\hat y}_{\theta_0,i}$ by the initial policy $\pi_{\theta_0}$ and obtain its improved response $\tilde{\mathbf{y}}_{\phi,i}$ from an external LLM $\pi_\phi$ with prompt $r$ such that
\begin{equation}\label{equation 2}
  \begin{aligned}
    \hat{\mathbf{y}}_{\theta_0,i} &\sim \pi_{\theta_0}\bigl(\cdot \mid \mathbf{x}_i\bigr),\\
    \tilde{\mathbf{y}}_{\phi,i} &\sim \pi_{\phi}\bigl(\cdot \mid r(\mathbf{x}_i, \mathbf{y}_i, \hat{\mathbf{y}}_{\theta_0,i})\bigr).
  \end{aligned}
\end{equation}

Then, we construct the training data as a set of $\{\mathbf{x}_i, \hat{\mathbf{y}}_{\theta_0,i},\tilde{\mathbf{y}}_{\phi,i}\}$ where $(\mathbf{x}_i, \hat{\mathbf{y}}_{\theta_0,i})$ and $\tilde{\mathbf{y}}_{\phi,i}$ act as an input and a target output, respectively. When we produce the target improved response $\tilde{\mathbf{y}}_{\phi,i}$, we use prompt $r$ to obtain the response for utilizing initial policy response $\hat{\mathbf{y}}_{\theta_0,i}$ and SFT response $\mathbf{y}_i$. To construct more online training data in terms of the policy response distribution, we compute the perplexity of each improver training data under our policy model and filter out outliers using the interquartile range (IQR) \cite{tukey1977exploratory} based filtering ${\mathcal F}$. Consequently, the resulting training dataset is defined as

\begin{equation}\label{equation 3}
\begin{aligned}
    \mathcal{D}_{R} &= \Bigl\{ \{(\mathbf{x}_i, \mathbf{\hat y}_{\theta_0,i}, \mathbf{\tilde y}_i)\}_{i=1}^{n} \,\big|\, (\mathbf{x}_i,\mathbf{\tilde y}_i) \in \mathcal{F}(\{(\mathbf{x}_i,\mathbf{\tilde y}_{\phi,i})\}_{i=1}^{n}) \cup \mathcal{F}(\{(\mathbf{x}_i,\mathbf{\hat y}_{\phi,i})\}_{i=1}^{n}) \Bigr\}, \\
    &\quad \quad \quad \quad \quad \quad \quad \quad \quad \quad \hat{\mathbf{y}}_{\phi,i} \sim \pi_{\phi}\bigl(\cdot \mid \mathbf{x}_i\bigr),
\end{aligned}
\end{equation}

where the response $\mathbf{\hat y}_{\phi,i}$ obtained by $\pi_{\phi}$ is included in the improver training data for more diverse targets for improvement. 

The objective function for training of the self-improver is now given as
\begin{equation}\label{equation 4}
{
\begin{aligned}
\displaystyle \theta^{*} = \arg\max_{\theta}\;
\mathbb{E}_{(\mathbf{x},\hat{\mathbf{y}}_{\theta_0},\tilde{\mathbf{y}})
  \sim \mathcal{D}_R}
\Bigl[
\log \pi_{\theta}\bigl(\tilde{\mathbf{y}} \mid r(\mathbf{x},\hat{\mathbf{y}}_{\theta_0})\bigr)
\Bigr],
\end{aligned}
}
\end{equation}

where the prompt $r$ serves as the input instruction for training the improver to refine the initial policy response $\hat{\mathbf{y}}_{\theta_0}$.
When training the improver, to prevent the improver from simply copying $y_i$ in generating the improved output, we exclude $y_i$ from the input condition of the improver. It is noted that the external LLM $\mathbf{\pi_\phi}$ is only used for constructing the training data for the improver, and the improver is trained only once with relatively small data before the successive preference optimizations.
Therefore, from the perspective of the preference optimization, this is the self-improving model using only on-policy self-generated preference data. In the following, we present the details of our proposed on-policy self-improver training procedure. First, section 3.1.1 introduces the prompting strategy to obtain target improved responses. Next, section 3.1.2 describes how we filter the target improved responses, and section 3.1.3 discusses the improver training method based on the filtered improved responses. Finally, section 3.1.4 presents the integrated improver-policy model designed for on-policy training.

\begin{table}[t]
  \caption{Win Rate (WR) comparison of improved response against initial policy response (response-B) and SFT response (response-A) on UltraFeedback and DPO-Mix-7K datasets. This evaluation measures how well the improved response aligns with the objective of the prompting setup designed for the external LLM (GPT-4 Turbo \cite{openai_gpt4turbo}) to generate a target improved response.}
  \label{Win Rate (WR) comparison of improved response}
  \centering
    \begin{tabular}{lcc}
      \toprule
      \textbf{Dataset} & \textbf{Comparison} & \textbf{WR (\%)} \\
      \midrule
      \multirow{2}{*}{\textbf{UltraFeedback}}
        & improved\_response vs response-B & 87.00 \\
        & improved\_response vs response-A & 50.50 \\
      \specialrule{0.4pt}{2pt}{2pt}
      \multirow{2}{*}{\textbf{DPO-Mix-7K}}
        & improved\_response vs response-B & 86.50 \\
        & improved\_response vs response-A & 45.00 \\
      \bottomrule
    \end{tabular}
\end{table} 

\subsubsection{Prompting for obtaining target Improved Responses}
For simplicity, we omit the subscript $i$ in $\tilde{\mathbf{y}}_{\phi, i}$ as $\mathbf{\tilde y_\phi}$, where each sample is drawn as $\tilde{\mathbf{y}}_{\phi, i} \sim \pi_{\phi}\bigl(\cdot \mid r(\mathbf{x}_i, \mathbf{y}_i, \hat{\mathbf{y}}_{\theta_{0},i})\bigr)$ and used as a target improved response for training the improver. The prompt $r$ in Eq.~(\ref{equation 2}) includes not only the initial policy response but also the SFT response $\mathbf{y}$ as high-quality reference response. In other words, the initial policy response, which requires improvement, is explicitly marked as a low-quality response (response-B), and the SFT response is marked as a high-quality response (response-A). Along with these data, we impose two constraints on the prompting to obtain a more effective target improved response from an external LLM (GPT-4 Turbo \cite{openai_gpt4turbo}).

First, to ensure that the improved response sufficiently reflects the characteristics of the initial policy response, we explicitly incorporate a constraint based on edit distance \cite{yoon2024tlcr}. This approach aims to generate the improved response $\mathbf{\tilde y_\phi}$ by encouraging the improved response to remain close to the initial policy response $\hat{\mathbf{y}}_{\theta_{0}}$ in terms of the edit distance. The ultimate goal of the improver is to learn how to enhance the initial policy response. To ensure the feasibility of this objective and to construct a dataset that provides useful guidance, the improved responses should not deviate significantly from the outputs of the initial policy and should be incrementally better while remaining closely aligned in distribution. 

Second, we constrain the model to generate a response with quality levels that are less than or equal to that of SFT response $\mathbf{y}$. This setting ensures that $\mathbf{y}$ acts as a reference for clear improving direction, while constraining the improved response to remain not too different from the initial policy response.

\begin{figure*}[h]
  \centering
  \includegraphics[width=0.65\linewidth]{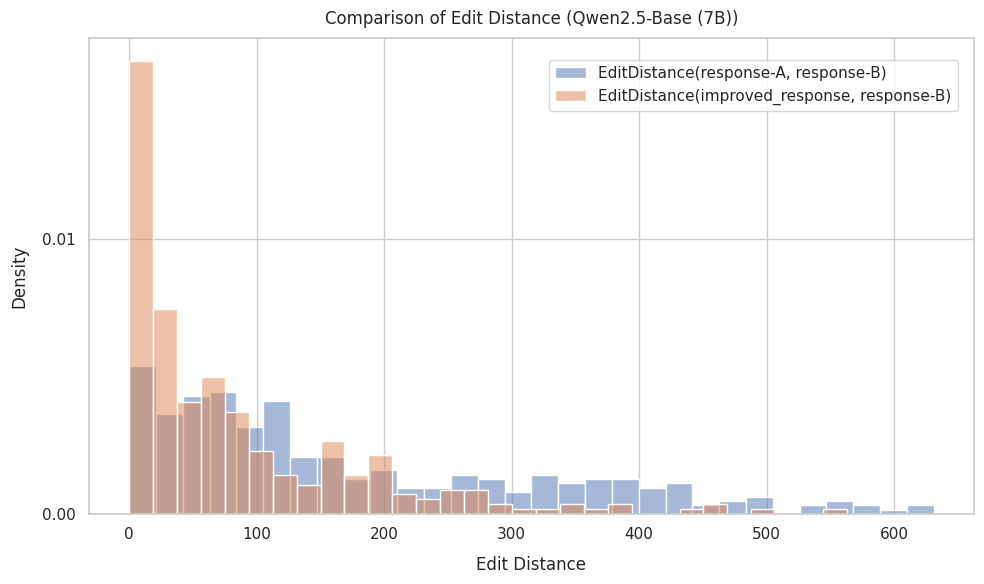}
  \caption {Histogram of edit distances comparing the improved response and the SFT response (response-A) against the initial policy response (response-B). Response-B is obtained from the Qwen2.5-Base (7B) model \cite{team2024qwen2}, and the improved response is extracted from the external LLM (GPT-4 Turbo \cite{openai_gpt4turbo}). The comparison is performed to verify whether the improved response satisfies the edit distance constraint defined in the prompting setup.}
  \label{edit_distance_target_improved_response1}
  \Description{Figure 3 shows the distribution of edit distance values between improved response and response-B, with a leftward shift compared to the comparison. This indicates that the edit distance values for improved response and response-B are generally smaller, meaning that improved response is more similar to response-B.}
\end{figure*}

\begin{figure*}[h]
  \centering
  \includegraphics[width=0.65\linewidth]{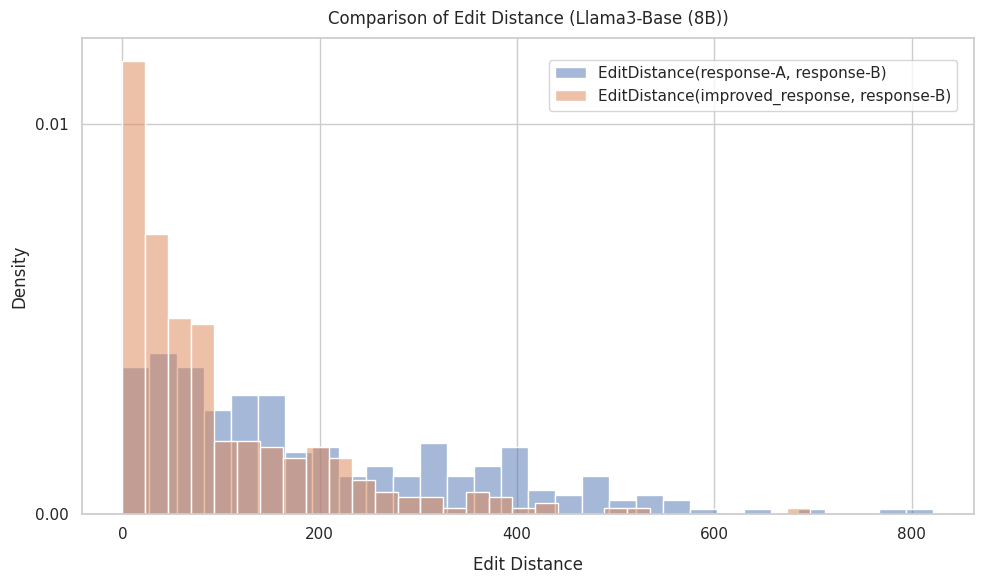}
  \caption {Histogram of edit distances comparing the improved response and the SFT response (response-A) against the initial policy response (response-B). Response-B is obtained from the Llama3-Base (8B) model \cite{grattafiori2024llama}, and the improved response is extracted from the external LLM (GPT-4 Turbo \cite{openai_gpt4turbo}).}
  \label{edit_distance_target_improved_response}
  \Description{Figure 4 shows the distribution of edit distance values between improved response and response-B, with a leftward shift compared to the comparison. This indicates that the edit distance values for improved response and response-B are generally smaller, meaning that improved response is more similar to response-B.}
\end{figure*} 

\begin{figure}[h]
    \centering
    Perplexity Histogram of Improved Response \par\medskip\medskip
    \begin{subfigure}[b]{0.485\textwidth}
        \includegraphics[width=\linewidth]{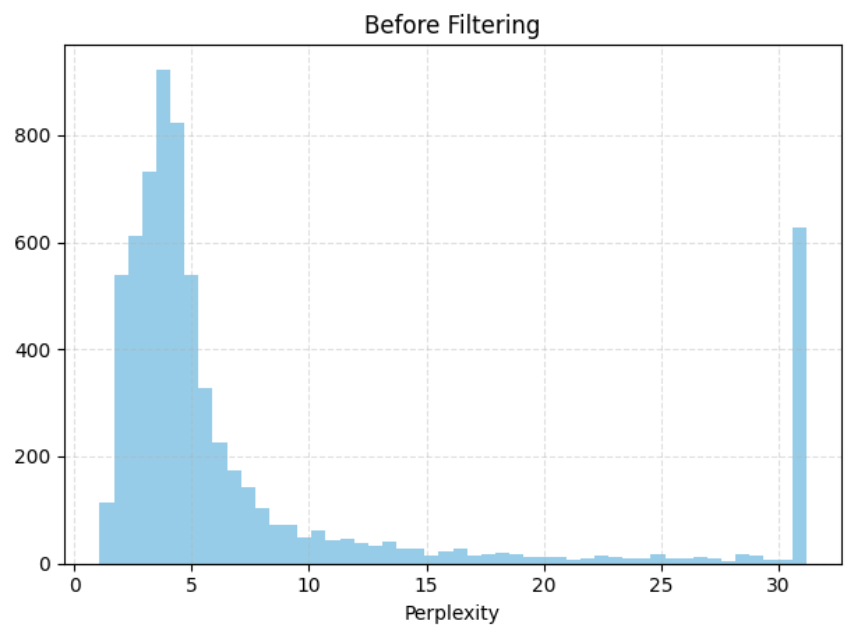}
        \label{fig:perplexity_histogram_of_improver_training_data_1}
    \end{subfigure}
    \begin{subfigure}[b]{0.45299\textwidth}
        \includegraphics[width=\linewidth]{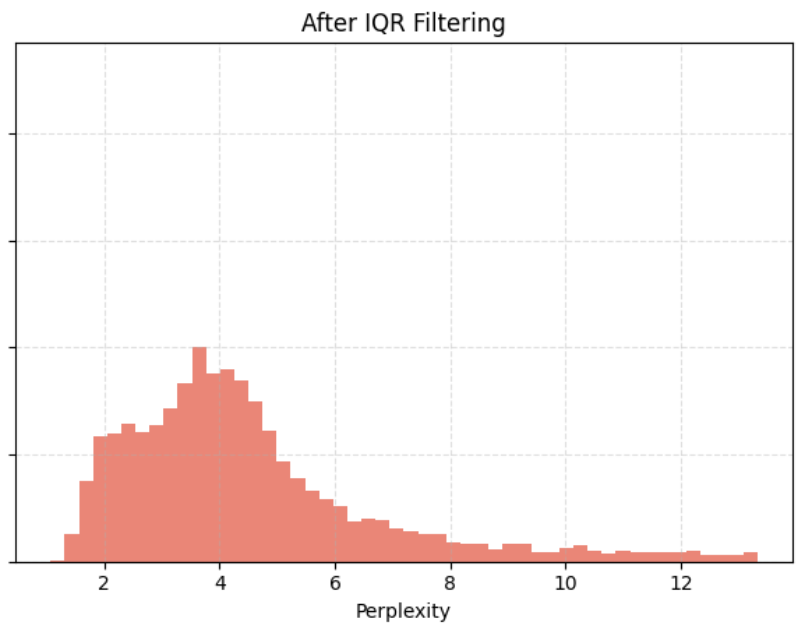}
        \label{fig:perplexity_histogram_of_improver_training_data_2}
    \end{subfigure}
    \medskip
    Perplexity Histogram of external LLM response \par\medskip\medskip
    \begin{subfigure}[b]{0.485\textwidth}
        \includegraphics[width=\linewidth]{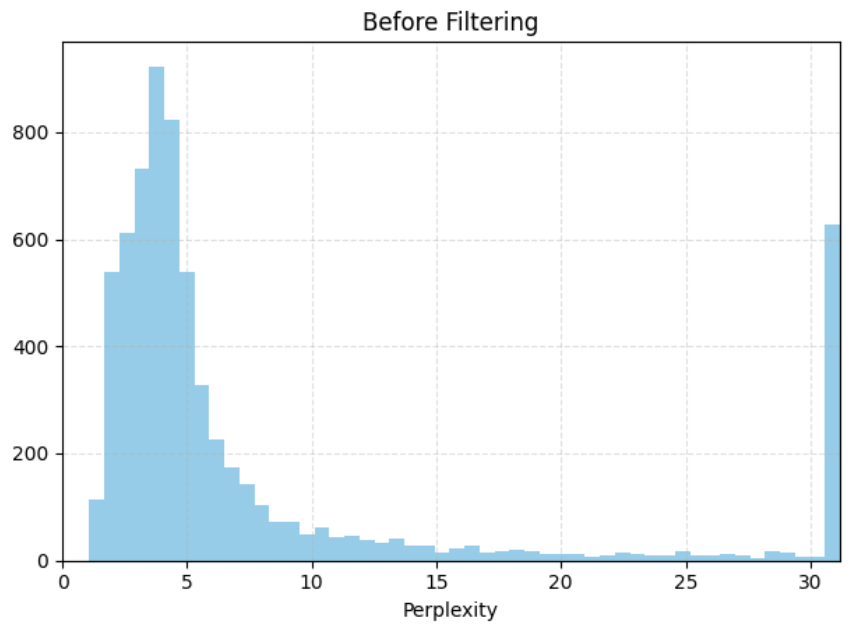}
        \label{fig:perplexity_histogram_of_improver_training_data_3}
    \end{subfigure}
    \begin{subfigure}[b]{0.4569\textwidth}
        \includegraphics[width=\linewidth]{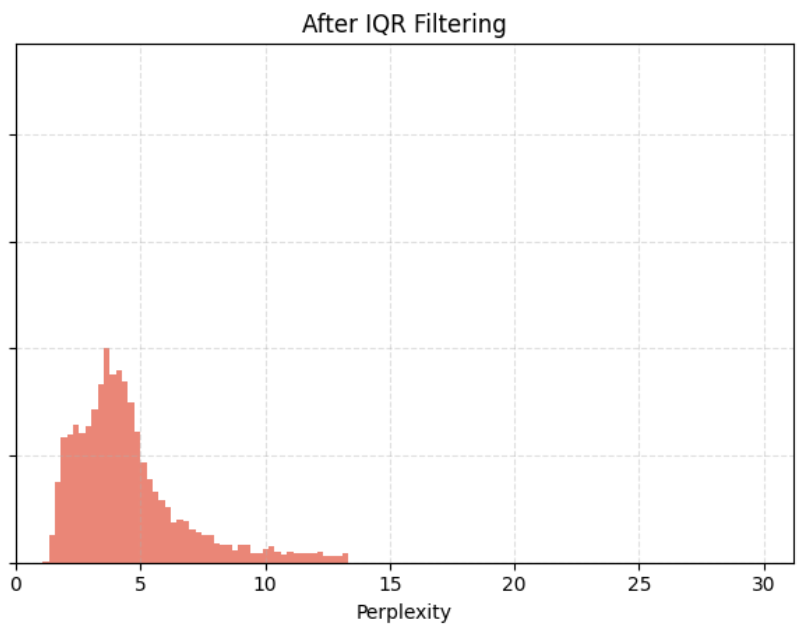}
        \label{fig:perplexity_histogram_of_improver_training_data_4}
    \end{subfigure}
    \caption{Histograms of perplexities of improved responses before and after IQR filtering (outliers removed). The top plot corresponds to the improved responses $\mathbf{\tilde y_\phi}$, while the bottom plot shows the external LLM responses $\mathbf{\hat y}_{\phi}$.}
    \Description{Figure 5 shows the distribution of perplexity values for improver data before and after filtering. When comparing the differences before and after filtering, it can be seen that outliers are removed after filtering clearly.}
    \label{fig:perplexity_histogram_of_improver_training_data}
\end{figure}

The final prompt used to generate the target improved response $\mathbf{\tilde y_\phi}$, which satisfies both constraints described above, is described in Figure~\ref{fig5_prompt2}. To verify whether the generated response $\mathbf{\tilde y_\phi}$ satisfies the intended prompt conditions, we conduct pairwise comparisons against the SFT response (response-A) and the initial policy response (response-B) using GPT-4 Turbo \cite{openai_gpt4turbo} as an automatic annotator. The win rate is then computed as the proportion of times the improved response is selected as better, and the corresponding results are presented in Table~\ref{Win Rate (WR) comparison of improved response}. In these results, we observe consistent trends on both datasets, UltraFeedback Cleaned \cite{cui2023ultrafeedback, allenai-ultrafeedback-binarized-cleaned} and DPO-Mix-7K \cite{dpo_mix_7k}. First of all, the metric \textbf{improved\_response vs response-B} evaluates whether the improved responses we extract show meaningful improvements over the initial policy responses (response-B), i.e., whether the improved responses consistently outperform response-B in quality. The metric \textbf{improved\_response vs response-A} is used to assess whether the improved responses exhibit a win rate that is similar to or lower than response-A. The results indicate that the improved responses achieve approximately 80\% higher quality compared to the initial policy responses while compared to the SFT responses, the improvement is about 50\%. In addition, it is important to assess whether the target improved response is generated while satisfying the edit distance constraint, which encourages the response to remain within a reasonable distance of the initial policy output distribution. We randomly sample 300 examples and compute the edit distances between the initial policy responses and the improved responses. For comparison, we also include the distances from the SFT response, and present the results in Figure~\ref{edit_distance_target_improved_response1}, \ref{edit_distance_target_improved_response}. The results show that the edit distances between the improved responses and the response-B are significantly lower than that between the improved responses and the response-A, indicating that the refinement process yields outputs that are more closely aligned with the initial policy responses. 
These results demonstrate that the improved responses successfully adhere to the constraints defined in the prompt.

\subsubsection{Filtering the target Improved Responses}
To construct training data for the improver that better aligns with the distribution of our target model, we apply a perplexity based filtering strategy. Given the set of improved response $\mathbf{\tilde y_\phi}$ and the set of response $\mathbf{\hat y}_{\phi}$ from the external LLM $\pi_{\phi}$, we filter the samples based on their alignment with the output distribution of the initial policy model $\pi_{\theta_{0}}$. Specifically, we aim to remove outliers—i.e., responses that significantly deviate from the distribution of $\pi_{\theta_{0}}$. To this end, we compute the perplexity of each sample using $\pi_{\theta_{0}}$ and identify outliers using the interquartile range (IQR) \cite{tukey1977exploratory} method: the lower quartile (Q1) and upper quartile (Q3) of the perplexity values, corresponding to the 25th and 75th percentiles respectively, are calculated, and the IQR is defined as $Q3 - Q1$. The lower and upper bounds are then set to $Q1 - 1.5 \times \mathrm{IQR}$ and $Q3 + 1.5 \times \mathrm{IQR}$, respectively. Samples with perplexity values outside this range are considered outliers and excluded from the training set. This statistical filtering allows us to retain only the most representative and meaningful data for effective improver training. Figure~\ref{fig:perplexity_histogram_of_improver_training_data} presents histograms of perplexity values before and after filtering. After IQR based filtering, outliers are removed, resulting in a more stable distribution concentrated around lower perplexity values. The final improver training data obtained after filtering is computed as follows: $\mathcal{D}_{R} = \bigr \{ \{(\mathbf{x}_i, \mathbf{\hat y}_{\theta_0,i}, \mathbf{\tilde y}_i)\}_{i=1}^{n}\mid (\mathbf{x}_i,\mathbf{\tilde y}_i) \in \mathcal{F}(\{(\mathbf{x}_i,\mathbf{\tilde y}_{\phi,i})\}_{i=1}^{n}) \cup \mathcal{F}(\{(\mathbf{x}_i,\mathbf{\hat y}_{\phi,i})\}_{i=1}^{n}) \bigr \}$.

\subsubsection{Improver Training}
\begin{figure} [t]
    \centering
    {\includegraphics[width=0.9\linewidth]{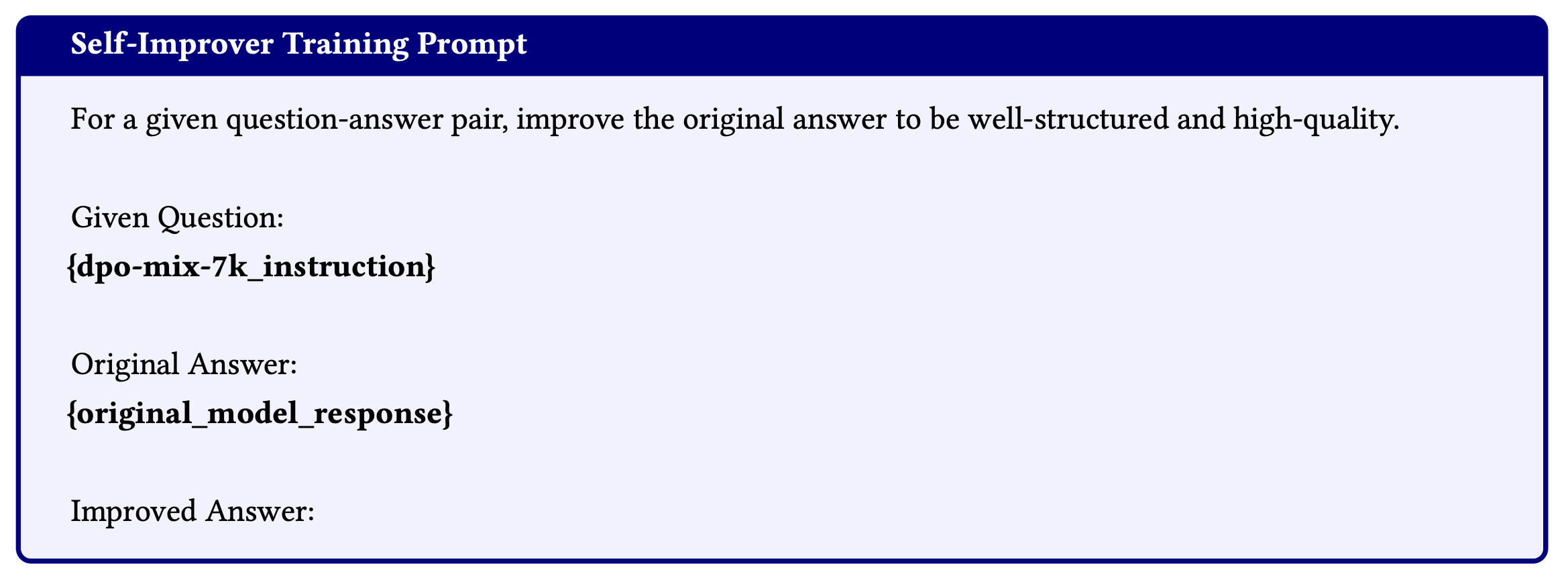}}
    \caption {Prompt $r$ for training self-improver, as denoted in Eq. (\ref{equation 4}).}
    \label{fig4_appendix}
    \Description{Figure 6. Fully described in the text.}
\end{figure}

To train the current policy model to serve as an improver, we adopt a dedicated objective tailored for response refinement. The improver training dataset $\{(\mathbf{x}, \mathbf{\hat y}_{\theta_0}), \mathbf{\tilde y}\}$ is constructed by pairing each instruction $x$ with the corresponding initial policy response $\mathbf{\hat y}_{\theta_0}$ as input, and setting the filtered improved response $\mathbf{\tilde y}$ as the target. The prompt $r$ in Eq.~(\ref{equation 4}) used for this training is described in detail in Figure~\ref{fig4_appendix}. This enables the model to learn how to refine its own outputs effectively.

\begin{figure}[t]
    \centering
    {\includegraphics[width=0.9\linewidth]{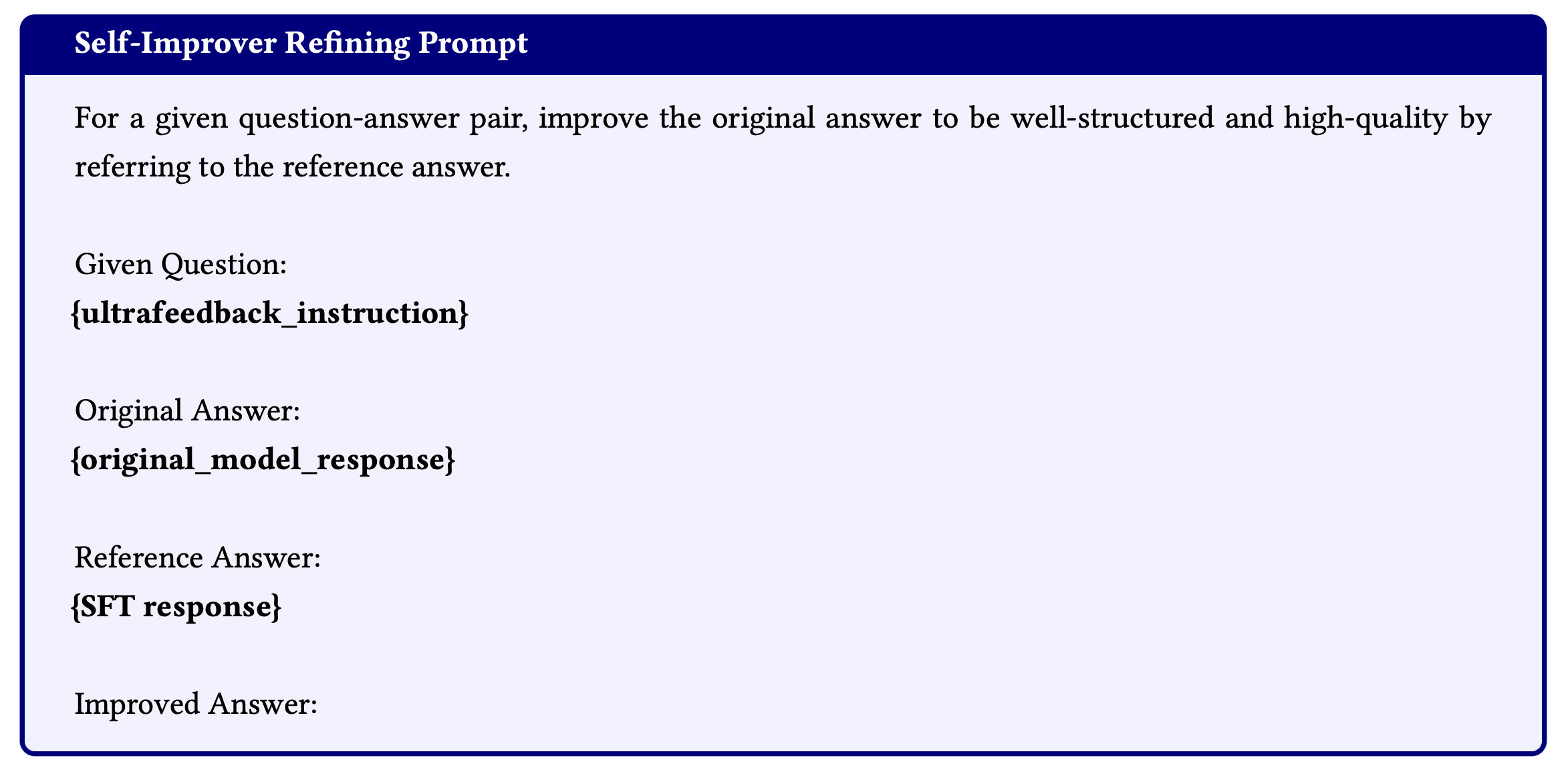}}
    \caption {Prompt $r$ for refining the response using self-improver, as denoted in Eq.~(\ref{equation 1}).}
    \label{fig6_appendix}
    \Description{Figure 7. Fully described in the text.}
\end{figure}

\subsubsection{Integrated Improver-Policy Model for On-Policy}
SGPO integrates the roles of the improver and the policy into a single model, enabling both roles through the prompt control. 
In contrast, the previous approach \cite{dong2024self} separates the improver and the policy, which has two main drawbacks:
improver cannot directly understand the generation mechanism of the current policy, and it is necessary to continuously maintain two different models during training.
To overcome these limitations, we adopt an integrated model. 
This approach allows the improver to understand the generation mechanism of the current policy, enabling more on-policy preference optimization throughout the process. For this in the following the optimized improver ${\pi_{\theta^*}}$ in Eq. \eqref{equation 4} is also used as a current policy to be updated.

\subsection{Self-Generated Preference Data and Optimization}
We generate responses $\mathbf{\hat y_{\theta^*}}$, $\mathbf{\tilde y_{\theta^*}}$ using the trained self-improver $\pi_{\theta^*}$, where the process for generating improved response $\mathbf{\tilde y_{\theta^*}}$ from the current $\mathbf{\hat y_{\theta^*}}$ response is defined as:
\begin{equation}\label{eq:4}
  \begin{aligned}
    \hat{\mathbf{y}}_{\theta^*} &\sim \pi_{\theta^*}\bigl(\cdot \mid \mathbf{x}\bigr),\\
    \tilde{\mathbf{y}}_{\theta^*} &\sim \pi_{\theta^*}\bigl(\cdot \mid r(\mathbf{x}, \mathbf{y}, \hat{\mathbf{y}}_{\theta^*})\bigr).
  \end{aligned}
\end{equation}
When the trained self-improver generates an improved response, we utilize SFT response $\mathbf{y}$ to guide the current policy.
The details of the refinement prompt used in this process are provided in Figure~\ref{fig6_appendix}. Then, we construct the on policy preference data $\mathcal{D}_P$ by assigning the improved response $\mathbf{\tilde y_{\theta^*}}$ as the chosen response and the current policy response $\mathbf{\hat y_{\theta^*}}$ as the rejected response such that $\mathcal{D}_P=\bigl\{(\mathbf{x}, {\mathbf{\tilde y}_{\theta^*}}, {\mathbf{\hat y}_{\theta^*}}) \bigr\}$. The final optimization objective function is defined as
\begin{equation}\label{eq:2}
\begin{aligned}
\theta^{**} = \arg\min_{\theta}\; \mathbb{E}_{(\mathbf{x},\tilde{\mathbf{y}}_{\theta^*},\hat{\mathbf{y}}_{\theta^*})
  \sim \mathcal{D}_P}\Bigl[
-\log \sigma\Bigl(
  \beta \log \frac{\pi_{\theta}(\tilde{\mathbf{y}}_{\theta^*}\mid \mathbf{x})}
               {\pi_{\theta^*}(\tilde{\mathbf{y}}_{\theta^*}\mid \mathbf{x})}
  -
  \beta \log \frac{\pi_{\theta}(\hat{\mathbf{y}}_{\theta^*}\mid \mathbf{x})}
               {\pi_{\theta^*}(\hat{\mathbf{y}}_{\theta^*}\mid \mathbf{x})}
\Bigr)
\Bigr],
\end{aligned}
\end{equation}
where $\sigma$ is the sigmoid function and $\beta$ is a parameter controlling the deviation from the reference policy $\pi_{\theta^*}$.

The overall algorithm of SGPO is described in Algorithm~\ref{alg:overview}. 

\begin{algorithm*} [t]
\caption{Self-Generated Preference Optimization (SGPO)}\label{alg:overview}
\begin{algorithmic}
  \STATE \textbf{Input:}
  \STATE \quad initial policy $\pi_{\theta_{0}}$, external LLM $\pi_{\phi}$, refine prompt $r$,
  \STATE \quad SFT data for improver training $S_R=\{(\mathbf{x}_{i}, \mathbf{y}_{i})\}_{i=1}^{n}$,
  \STATE \quad SFT data for policy update $S_P= \{(\mathbf{x}_{j},\mathbf{y}_{j})\}_{j=1}^{m}$,
  \STATE \quad data filter $\mathcal{F}$, improver training data $\mathcal{D}_{R}=\emptyset$, synthetic preference data $\mathcal{D}_{P}=\emptyset$.
  \STATE \textbf{Step 1: Improver Training}
  \STATE Generate policy responses by $\pi_{\theta_{0}}$: $\mathbf{\hat y}_{\theta_0,i} \sim \pi_{\theta_{0}}(\cdot \mid \mathbf{x}_{i})$, $\forall i\in \{1,..,n\}$
  \STATE Generate improved responses by $\pi_{\phi}$: $\mathbf{\tilde y}_{\phi,i} \sim \pi_{\phi}(\cdot \mid r(\mathbf{x}_{i}, \mathbf{y}_{i}, \mathbf{\hat y}_{\theta_0,i}))$, $\forall i\in \{1,..,n\}$
  \STATE Generate policy responses by $\pi_{\phi}$: $\mathbf{\hat y}_{\phi,i} \sim \pi_{\phi}(\cdot \mid \mathbf{x}_{i})$, $\forall i\in \{1,..,n\}$
  \STATE Filter out improved responses and external LLM responses, and construct improver training data:
  \STATE \quad
    $\displaystyle
      \mathcal{D}_{R}= \left\{ (\mathbf{x}_i, \mathbf{\hat y}_{\theta_0,i}, \mathbf{\tilde y}_i), \forall i=\{1,...,n\} \mid (\mathbf{x}_i,\mathbf{\tilde y}_i) \in \mathcal{F}(\{(\mathbf{x}_i,\mathbf{\tilde y}_{\phi,i})\}_{i=1}^{n}) \cup \mathcal{F}(\{(\mathbf{x}_i,\mathbf{\hat y}_{\phi,i})\}_{i=1}^{n}) \right\}$
  \STATE Train $\pi_{\theta_0}$ using $\mathcal{D}_{R}$:
  \STATE \quad
    $\displaystyle \theta^{*} = \argmax_{\theta}\;
      \mathbb{E}_{(\mathbf{x}_i,\mathbf{\hat y}_{\theta_0,i},\tilde{\mathbf{y}}_i) \sim \mathcal{D}_R}
      \left[\log \pi_{\theta}(\tilde{\mathbf{y}}_i \mid r(\mathbf{x}_i,\mathbf{\hat y}_{\theta_0,i}))\right]$
  \STATE \textbf{Step 2: Self-Generated Preference Data and Optimization}
  \STATE Generate policy responses by $\pi_{\theta^{{*}}}$: $\mathbf{\hat y}_{\theta_{j}^{*}} \sim \pi_{\theta^{{*}}}(\cdot \mid \mathbf{x}_{j})$, $\forall j\in \{1,...,m\}$
  \STATE Generate improved responses by $\pi_{\theta^{{*}}}$ : ${\mathbf{\tilde y}_{\theta^*_j}} \sim \pi_{\theta^{*}}(\cdot \mid r(\mathbf{x}_{j}, \mathbf{y}_{j}, \mathbf{\hat y}_{\theta_{j}^{*}}))$, $\forall j\in \{1,...,m\}$
  \STATE Construct preference data $\mathcal{D}_P=\left\{(\mathbf{x}_j, {\mathbf{\tilde y}_{\theta^*_j}},{\mathbf{\hat y}_{\theta^*_j}}) \right\}_{j=1}^m$ and Optimize:
  \STATE \quad
    $\displaystyle \theta^{**} = \argmin_{\theta}\;
      \mathbb{E}_{(\mathbf{x}_{j}, {\mathbf{\tilde y}_{\theta^*_j}},{\mathbf{\hat y}_{\theta^*_j}})\sim \mathcal{D}_P}
      \left[
        -\log\sigma\left(
          \beta\log\frac{\pi_{\theta}({\mathbf{\tilde y}_{\theta^*_j}}\mid \mathbf{x}_{j})}
                       {\pi_{\theta^{*}}({\mathbf{\tilde y}_{\theta^*_j}}\mid \mathbf{x}_{j})}
          -\beta\log\frac{\pi_{\theta}({\mathbf{\hat y}_{\theta^*_j}}\mid \mathbf{x}_{j})}
                       {\pi_{\theta^{*}}({\mathbf{\hat y}_{\theta^*_j}}\mid \mathbf{x}_{j})}
        \right)
      \right]$
\end{algorithmic}
\end{algorithm*}

\section{Experiments}

\subsection{Datasets}
We use the SFT split of the UltraChat \cite{ding2023enhancing} dataset for training an SFT model as our initial policy model $\pi_{\theta_0}$. 
UltraChat is a refined conversational dataset comprising 200K dialogues generated by ChatGPT \cite{ouyang2022training}, covering diverse subject areas and multiple conversational styles. In all training settings where model responses are used as data, we employ vLLM \cite{kwon2023efficient} with a sampling-based decoding strategy for inference. We set a sampling temperature to 0.7 and top-p to 0.9 for all generations. 

\subsubsection{Improver Training}
For training the response improver, we employ the SFT split of the Dpo-Mix-7K dataset \cite{dpo_mix_7k}, which comprises high-quality chosen responses. The chosen responses are used as references to obtain the target improved responses for training the improver. From this data, we use an external LLM (GPT-4 Turbo \cite{openai_gpt4turbo}) to generate the target improved responses and external LLM responses. After conducting perplexity based IQR filtering on each of the 7K datasets, the final number of instances utilized in our experiments is 13,092 for Qwen2.5-Base (7B) \cite{team2024qwen2}, 11,561 for Llama3-Base (8B) \cite{grattafiori2024llama}, and 12,887 for Qwen2-Base (1.5B) \cite{yang2024qwen2technicalreport}.

\subsubsection{Policy Finetuning}
For SGPO preference optimization, we use the UltraFeedback Cleaned dataset \cite{cui2023ultrafeedback, allenai-ultrafeedback-binarized-cleaned}, which consists of 60.8K prompts and is a pre-processed version of the original UltraFeedback dataset. The original UltraFeedback dataset consists of 64K prompts, where each prompt is accompanied by four model completions. GPT-4 \cite{achiam2023gpt} is used to assign a score to each completion, along criteria like helpfulness and honesty. Here, it is noted that in our training setting, we do not use preference labels in the UltraFeedback dataset. We utilize only the SFT split, which contains prompt-chosen response pairs for policy finetuning.

\subsection{Models and Training Setup}
We conduct experiments using three different base LLMs: Qwen2.5-Base (7B) \cite{team2024qwen2}, Llama3-Base (8B) \cite{grattafiori2024llama}, and Qwen2-Base (1.5B) \cite{yang2024qwen2technicalreport}.
All models are trained with an effective batch size of 128, a maximum sequence length of 2048, and a cosine learning rate scheduler.
For supervised fine-tuning on the UltraChat-200K dataset, Qwen2.5-Base (7B) and Llama3-Base (8B) are trained for 1 epoch with a learning rate of 2e-5 while Qwen2-Base (1.5B) is trained for 3 epochs with a learning rate of 2e-5. 

\subsubsection{Baseline Settings}
For the DPO baseline \cite{rafailov2023direct}, we train Qwen2.5-Base (7B) and Llama3-Base (8B) on the UltraFeedback preference dataset for 1 epoch with a learning rate of 5e-7 and $\beta$ = 0.01.
For the SPIN baseline \cite{chen2024self}, we set a parameter $\beta = 0.5$ and use a learning rate of 3e-7 for Qwen2.5-Base (7B) and 5e-7 for Llama3-Base (8B). Each training preference pair is constructed using the chosen response from the UltraFeedback dataset as positive example and the initial policy response as negative example.

\subsubsection{SGPO Settings}
SGPO training consists of two sequential steps: In Step 1 (Improver Training), we train the model to act as a self-improver for 3 epochs using the same hyperparameters as in the supervised fine-tuning phase. In Step 2 (Preference Optimization), we construct synthetic preference pairs $\mathcal{D}_{P}$ by pairing the responses from the policy with their improved versions generated by the self-improver.
We train the model for 1 epoch with a learning rate of 5e-7 and $\beta = 0.5$. 

\subsubsection{SGPO Variants}
As discussed in Section~\ref{sec:Method}, a key design choice in SGPO is the parameter sharing between the policy and the improver to ensure that the refinement process is more policy-aware. Also, the improver in SGPO is trained with target improved responses based on the incremental response improvement.
To validate this design choice, we include the following two SGPO variants:

\paragraph{SGPO$^{\dagger}$ (No parameter sharing)} This variant removes the shared architecture: the improver and policy are trained as separate models. The role of the improver is restricted to improving the responses produced by the policy model. As a result, self-improvement is no longer performed by the policy itself.

\paragraph{SGPO$^{\ddagger}$ (No improved supervision)} In this variant, the improver is trained without access to our proposed improved responses $\mathbf{\tilde y_\phi}$. Unlike the SGPO setting where both external LLM responses and improved responses are used to train the self-improver, this variant trains the improver using only external LLM responses. In addition, the shared architecture is also removed in this setting, so the improver and policy are trained separately. Note that this variant resembles SynPO \cite{dong2024self} in that the response refinement is conducted by a separate improver, and the improver is trained using only external LLM responses.

\begin{table*}[t]
  \caption{Evaluation details for AlpacaEval 2.0 and Arena-Hard. The baseline model denotes the model used for comparative evaluation. GPT-4 Turbo corresponds to GPT-4-Preview-1106.}
  \label{tb1}
  \centering
  \resizebox{\textwidth}{!}{
    \begin{tabular}{lccccc}
      \toprule
      \textbf{} & \textbf{\# Exs.} & \textbf{Baseline Model} & \textbf{Judge Model} & \textbf{Scoring Type} & \textbf{Metric} \\
      \midrule
      AlpacaEval 2.0 & 805 & GPT-4 Turbo & GPT-4 Turbo & Pairwise comparison & LC \& raw win rate \\
      Arena-Hard & 500 & GPT-4-0314 & GPT-4 Turbo &  Pairwise comparison & Win rate \\
      \bottomrule
    \end{tabular}
  }
\end{table*}

\subsection{Evaluation Benchmarks} 
We evaluate our approach on two widely adopted benchmarks for LLM alignment, as summarized in Table~\ref{tb1}: AlpacaEval 2.0 \cite{alpaca_eval} and Arena-Hard \cite{arenahard2024, li2024crowdsourced}. 
These benchmarks assess models' conversational capabilities across a diverse range of prompts and are commonly used to measure the quality of instruction following models. 

\begin{table*}[t]
  \caption{Evaluation results on AlpacaEval 2.0 and Arena-Hard. LC(\%) and WR(\%) refer to Length-Controlled Win Rate and Win Rate, respectively. Performance comparison is conducted with Qwen2.5-Base (7B) and Llama3-Base (8B). SGPO$^{\dagger}$ and SGPO$^{\ddagger}$ indicate variant settings of SGPO.} 
  \label{tb2}
  \centering
  \resizebox{\textwidth}{!}{
  \begin{tabular}{l ccc ccc ccc}
    \toprule
    \multirow{3}{*}{\textbf{Method}}
      & \multicolumn{3}{c}{\textbf{Qwen2.5-Base (7B)}}
      & \multicolumn{3}{c}{\textbf{Llama3-Base (8B)}} \\
    \cmidrule(lr){2-4} \cmidrule(lr){5-7} \cmidrule(lr){8-10}
      & \multicolumn{2}{c}{\textbf{AlpacaEval 2.0}} & \textbf{Arena-Hard}
      & \multicolumn{2}{c}{\textbf{AlpacaEval 2.0}} & \textbf{Arena-Hard} \\
    \cmidrule(lr){2-3} \cmidrule(lr){4-4}
    \cmidrule(lr){5-6} \cmidrule(lr){7-7}
      & \textbf{LC (\%)} & \textbf{WR (\%)} & \textbf{WR (\%)} 
      & \textbf{LC (\%)} & \textbf{WR (\%)} & \textbf{WR (\%)} \\
    \midrule
    SFT & 5.8 & 4.0 & 7.7    & 7.3    & 4.7    & 4.9        \\
    DPO & 9.2 & 9.9 & 23.9    & 14.3    & 12.3    & 12.7        \\
    SPIN & 5.5    & 4.8    & 8.6    & 7.5    & 5.8    & 3.4        \\
    SGPO$^{\dagger}$ & 6.6 & 5.5 & 7.9  & 9.0    & 7.4    & 7.0        \\
    SGPO$^{\ddagger}$ & 11.6 & 7.5 & 18.3  & 9.4    & 8.5    & 5.7        \\
    SGPO & \textbf{25.2} & \textbf{29.6} &\textbf{41.2}  & \textbf{20.6}    & \textbf{26.2}    & \textbf{21.8}       \\
    \bottomrule
  \end{tabular}
  }
\end{table*}

\begin{figure}[h]
  \centering
  \includegraphics[width=0.5\linewidth]{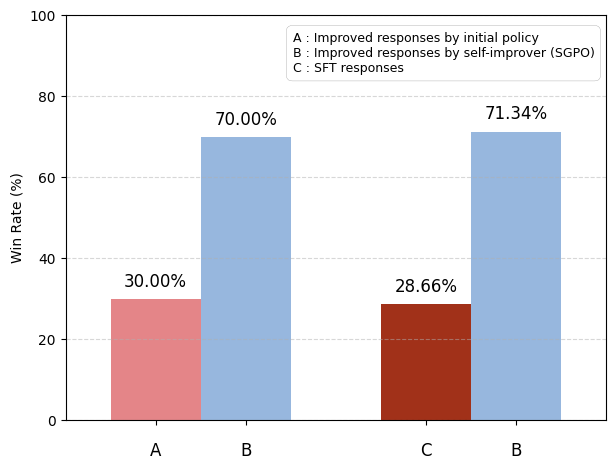}
  \caption {Performance evaluation of the self-improver. A represents the improved responses $\tilde{\mathbf{y}}_{\theta_0}$ generated by the initial policy $\pi_{\theta_0}$ to improve a given current policy responses $\mathbf{\hat y_{\theta^*}}$ (i.e., $\tilde{\mathbf{y}}_{\theta_0} \sim \pi_{\theta_0}\bigl(\cdot \mid r(\mathbf{x}, \mathbf{y}, \hat{\mathbf{y}}_{\theta^*})\bigr)$). B shows the improved responses $\tilde{\mathbf{y}}_{\theta^*}$ generated by the self-improver $\pi_{\theta^*}$ to improve a given responses $\mathbf{\hat y_{\theta^*}}$ (i.e., $\tilde{\mathbf{y}}_{\theta^*} \sim \pi_{\theta^*}\bigl(\cdot \mid r(\mathbf{x}, \mathbf{y}, \hat{\mathbf{y}}_{\theta^*})\bigr)$). C indicates SFT responses $\mathbf{y}$.}
  \label{fig2}
  \Description{Figure 8. Fully described in the text.}
\end{figure}

\begin{figure}[h]
  \centering
  \includegraphics[width=0.5\linewidth]{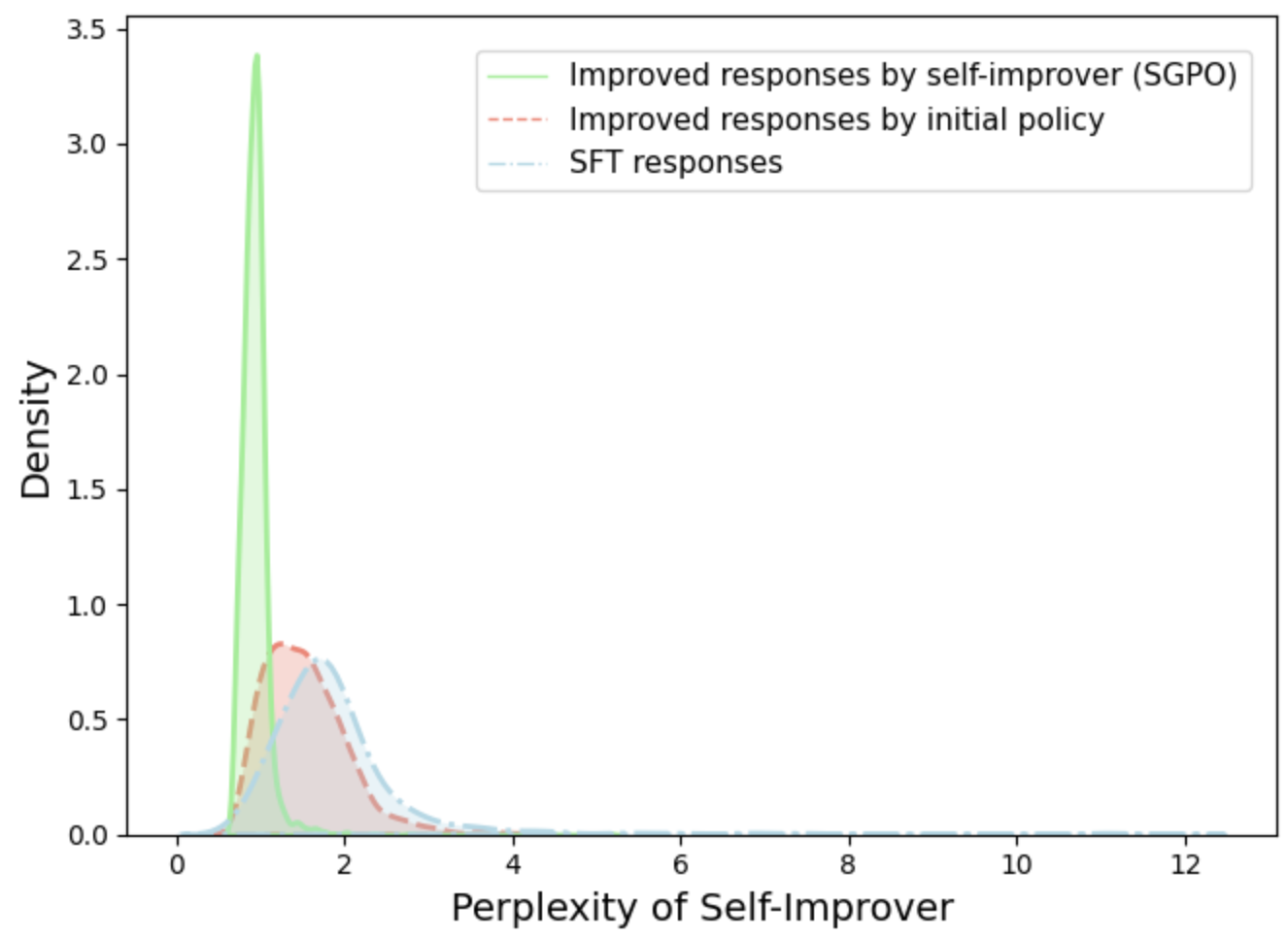}
  \caption {Histogram of log-scaled perplexity values measured by the self-improver for each set of responses.}
  \label{fig3}
  \Description{Figure 9 shows the perplexity distribution for each data type based on the self-improver criterion. The improved responses represented by a line showing the lowest values and narrowest spread. The other two datasets, represented by different dotted lines, have higher perplexity values than the improved responses.}
\end{figure}

AlpacaEval 2.0 employs GPT-4 Turbo \cite{achiam2023gpt} as an automated evaluator in a zero-shot pairwise comparison setup, covering 805 prompts. It reports two metrics: \textit{Length-Controlled Win Rate} (LC) and \textit{Win Rate} (WR). 
The LC metric mitigates the bias toward overly verbose response, offering a more balanced comparison.
Arena-Hard consists of 500 challenging prompts. 
Evaluation is performed using a GPT-4 automated judge via pairwise comparisons, and results are reported in terms of WR. 
We report WR as the percentage of wins against the baseline model. For AlpacaEval 2.0 evaluation, we use a sampling-based decoding strategy where a temperature is set to 0.7 and top-p is set to 0.9. For Arena-Hard, we use a default greedy decoding.

\subsection{Experiment Results}

\subsubsection{Main Results}
As shown in Table~\ref{tb2}, SGPO substantially improves performance and achieves the highest overall performance across both benchmarks AlpacaEval 2.0 and Arena-Hard, whereas most preference-based algorithms only modestly improve performance over SFT.
SGPO outperforms DPO by 6.28 to 16.18 points on AlpacaEval 2.0 in LC, and by 9.1 to 17.3 points on Arena-Hard in WR. Compared to SPIN, SGPO achieves consistently higher performance across all benchmarks. 
On Qwen2.5-Base (7B), SGPO improves the performance over SGPO$^{\dagger}$ and SGPO$^{\ddagger}$ by up to 18.66\% and 13.66\% on the AlpacaEval 2.0 LC, and by up to 33.3\% and 22.9\% on Arena-Hard, respectively. Under the Llama3-Base (8B) setting, SGPO also consistently outperforms both variants across all benchmark tasks. The performance drop observed in both SGPO$^{\dagger}$ and SGPO$^{\ddagger}$ underscores the contribution of our design choices to the effectiveness of the SGPO framework. These results demonstrate the effectiveness of using the policy model itself as the improver, trained with improved responses that remain within the distribution of the initial policy responses.

\subsubsection{Self-Improver Evaluation}
We evaluate the effectiveness of the improved responses generated by the self-improver from two perspectives, comparing the improved responses from the initial policy $\pi_{\theta_0}$ and the SFT responses $\mathbf{y}$ as baselines.
All evaluations are conducted using the Llama3-Base (8B) model on the UltraFeedback dataset.
Here, $\theta^{*}$ acts as both the improver and the policy after the Step 1 (Improver Training). All target responses for improvement are generated by $\pi_{\theta^*}$. 
Accordingly, the comparison is conducted among the following sets of responses: 
The improved responses generated by the self-improver as $\tilde{\mathbf{y}}_{\theta^*} \sim \pi_{\theta^*}\bigl(\cdot \mid r(\mathbf{x}, \mathbf{y}, \hat{\mathbf{y}}_{\theta^*})\bigr)$, the improved responses generated from the initial policy as $\tilde{\mathbf{y}}_{\theta_0} \sim \pi_{\theta_0}\bigl(\cdot \mid r(\mathbf{x}, \mathbf{y}, \hat{\mathbf{y}}_{\theta^*})\bigr)$, and the high-quality SFT responses $\mathbf{y}$.

First, we evaluate the quality of the self-improver in refining responses. Using GPT-4 Turbo as an automated annotator, we perform pairwise comparisons on 500 samples, comparing the improved responses $\mathbf{\tilde y_{\theta^*}}$ against each of the improved responses $\mathbf{\tilde y_{\theta_0}}$ and the SFT responses $\mathbf{y}$. The comparison focuses on which responses better refine the current policy responses $\mathbf{\hat y_{\theta^*}}$.
As shown in Figure~\ref{fig2}, the responses refined by the trained self-improver $\pi_{\theta^*}$ achieve a win rate of approximately 70\% when compared to both $\tilde{\mathbf{y}}_{\theta_0}$ and $\mathbf{y}$, demonstrating better refinement.

Second, we analyze the perplexity of each set of improved responses—$\mathbf{\tilde y_{\theta^*}}$, $\mathbf{\tilde y_{\theta_0}}$, and $\mathbf{y}$—under the model $\mathbf{\pi_{\theta^*}}$ to determine their alignment with the policy distribution.
Figure~\ref{fig3} shows the histogram of perplexity across 2,000 randomly selected samples for each response type, with $\mathbf{\tilde y_{\theta_0}}$ in red, $\mathbf{\tilde y_{\theta^*}}$ in green, and $\mathbf{y}$ in blue. 
Lower perplexity indicates that the improved responses are easier for the model to generate, reflecting better alignment onlineness with the policy distribution.
The results indicate that the improved responses $\mathbf{\tilde y_{\theta^*}}$ by the self-improver are more closely aligned with the probability distributions of the current policy to be updated by preference optimization. \\

\begin{table}[t]
  \caption{Comparison of preference pairs from on-policy and off-policy approaches. SGPO shows significant gains by an online preference dataset.}
  \label{tb4}
  \centering
    \begin{tabular}{lccccc}
      \toprule
      \multirow{2}{*}{\textbf{Method}}
        & \multirow{2}{*}{\textbf{Chosen}}
        & \multirow{2}{*}{\textbf{Rejected}}
        & \multicolumn{2}{c}{\textbf{AlpacaEval 2.0}} \\
      \cmidrule(lr){4-5}
       &  & 
        & \textbf{LC (\%)} & \textbf{WR (\%)} \\
      \midrule
      SPIN      & $\mathbf{y}$ & $\mathbf{\hat y_{\theta_0}}$ & 7.45 & 5.78 \\
      SGPO$^{\diamond}$ & $\mathbf{y}$ & $\mathbf{\hat y_{\theta^*}}$ & 15.00 & 10.23 \\
      SGPO      & $\mathbf{\tilde y_{\theta^*}}$ & $\mathbf{\hat y_{\theta^*}}$ & \textbf{20.62} & \textbf{26.21}  \\
      \bottomrule
    \end{tabular}
\end{table}

\begin{table*}[t]
  \caption{Evaluation results on AlpacaEval 2.0, focusing on the effects of improver training data and unification of the improver and the policy model.}
  \label{tb5}
  \centering
  \resizebox{\textwidth}{!}{%
    \begin{tabular}{l c c c c}
      \toprule
      \multirow{2}{*}{\textbf{Method}}
        & \multirow{2}{*}{\textbf{Improver Training Data $(\mathcal{D}_R)$}}
        & \multirow{2}{*}{\textbf{Integrating Improver-Policy}}
        & \multicolumn{2}{c}{\textbf{AlpacaEval 2.0}} \\
      \cmidrule(lr){4-5}
      & & & \textbf{LC (\%)} & \textbf{WR (\%)} \\
      \midrule
      SGPO-v1   & $\mathbf{\hat y_{\phi}}$ & \ding{55} & 3.10 & 3.30 \\
      SGPO-v2   & $\mathbf{\hat y_{\phi}}$ & \ding{51} & 3.41 & 4.42 \\
      SGPO-v3   & $\mathbf{\hat y_{\phi}}, \mathbf{\tilde y_{\phi}}$ & \ding{55} & 2.85 & 3.06 \\
      SGPO         & $\mathbf{\hat y_{\phi}}, \mathbf{\tilde y_{\phi}}$ & \ding{51} & \textbf{3.46} & \textbf{4.82} \\
      \bottomrule
    \end{tabular}%
  }
\end{table*}

\begin{table*}[t]
  \caption{Effectiveness of the self-improving loop without additional retraining of the improver. SGPO + Step 2 refers to applying the Step 2 (preference optimizaiton) one more time, using SGPO.}
  \label{tb6}
  \centering
  \resizebox{\textwidth}{!}{
  \begin{tabular}{l ccc ccc ccc}
    \toprule
    \multirow{3}{*}{\textbf{Method}}
      & \multicolumn{3}{c}{\textbf{Qwen2.5-Base (7B)}}
      & \multicolumn{3}{c}{\textbf{Llama3-Base (8B)}} \\
    \cmidrule(lr){2-4} \cmidrule(lr){5-7} \cmidrule(lr){8-10}
      & \multicolumn{2}{c}{\textbf{AlpacaEval 2.0}} & \textbf{Arena-Hard}
      & \multicolumn{2}{c}{\textbf{AlpacaEval 2.0}} & \textbf{Arena-Hard} \\
    \cmidrule(lr){2-3} \cmidrule(lr){4-4}
    \cmidrule(lr){5-6} \cmidrule(lr){7-7}
      & \textbf{LC (\%)} & \textbf{WR (\%)} & \textbf{WR (\%)} 
      & \textbf{LC (\%)} & \textbf{WR (\%)} & \textbf{WR (\%)} \\
    \midrule
    SFT           & 5.8  & 4.0  & 7.7 & 7.3 & 4.7 & 4.9 \\
    SGPO          & \textbf{25.2} & 29.6 & 41.2 & 20.6 & 26.2 & 21.8 \\
    SGPO + Step 2 & 23.6 & \textbf{34.2} & \textbf{44.6} & \textbf{22.4} & \textbf{34.5} & \textbf{21.9} \\
    \bottomrule
  \end{tabular}
  }
\end{table*}

\subsubsection{Ablation Studies} \label{subsubsec:AblationStudies}

\paragraph{Impact of On-Policy vs. Off-Policy Preference Dataset}
To evaluate the impact of on-policy data, we conduct experiments by constructing three distinct preference pair sets and perform the preference optimization on the Llama3-Base (8B) model.
As described in Table~\ref{tb4}, SPIN \cite{chen2024self} constructs preference data by using SFT response $\mathbf{y}$ as the chosen response and initial policy response $\mathbf{\hat y_{\theta_0}}$ as the rejected response; 
SGPO$^{\diamond}$ uses $\mathbf{y}$ as the chosen response and current policy response $\mathbf{\hat y_{\theta^*}}$ as the rejected response; SGPO uses improved response $\mathbf{\tilde y_{\theta^*}}$ as the chosen response and $\mathbf{\hat y_{\theta^*}}$ as the rejected response.
SGPO, in which both chosen and rejected responses satisfy the on-policy condition, achieves the best overall performance. Under the off-policy setting, both SPIN and SGPO$^{\diamond}$ exhibit lower performance. SGPO clearly demonstrates the beneficial effects of utilizing on-policy preference dataset.

\paragraph{Effectiveness of Improved Response and Model Integration} 
We investigate the impact of improver training data and model integration separately, as shown in Table~\ref{tb5}. 
The experiments are conducted using Qwen2-Base (1.5B). SGPO-v1 has a separated improver and policy model trained only an external LLM response $\mathbf{\hat y_\phi}$. SGPO-v2 maintains using the same external LLM response but unifies the improver and policy. The SGPO-v3 setting includes the improved response $\mathbf{\tilde y_\phi}$ in the training data but does not integrate improver and policy.
When using only an external LLM response $\mathbf{\hat y_\phi}$ as training data, SGPO-v2, which unifies the improver and policy, outperforms SGPO-v1, where the improver and policy are separated. SGPO-v3, which uses the same improver training data as SGPO but maintains a separated improver-policy structure, leads to the lowest performance. SGPO achieves the best performance by combining two key factors: using an integrated improver-policy model and incorporating the improved response $\mathbf{\tilde y_\phi}$ in the training data. This response is a progressively refined version of the initial policy model response. These results demonstrate that incorporating the improved response $\mathbf{\tilde y_\phi}$ and the self-improver mechanism, particularly the integrated improver-policy, is essential in producing beneficial preference data for preference optimization.

\paragraph{Self-Improving Loop Without Re-training the Improver}
To evaluate the additional self-boosting capability of SGPO, we perform an additional Step 2 of preference optimization using the trained policy as SGPO. 
This experiment is conducted with the Llama3-Base (8B) model. We denote this extended approach as SGPO + Step 2 in Table~\ref{tb6}. 
Let $\pi_{\theta^{**}}$ denote the policy fully optimized via SGPO. To perform the SGPO + Step 2 training, we extract policy response $\mathbf{\hat y}_{\theta^{**}} \sim \pi_{\theta^{**}}(\cdot \mid \mathbf{x})$ and improved response ${\mathbf{\tilde y}_{\theta^{**}}} \sim \pi_{\theta^{**}}(\cdot \mid r(\mathbf{x}, \mathbf{y}, \mathbf{\hat y}_{\theta^{**}}))$ by $\pi_{\theta^{**}}$. Then, we construct the preference data $\mathcal{D}_{P^{**}}$ by assigning the improved response ${\mathbf{\tilde y}_{\theta^{**}}}$ as the chosen and the policy response $\mathbf{\hat y}_{\theta^{**}}$ as the rejected such that $\mathcal{D}_{P^{**}}=\bigl\{(\mathbf{x}, \mathbf{\tilde y}_{\theta^{**}}, \mathbf{\hat y}_{\theta^{**}}) \bigr\}$. The final optimization objective function is defined as 
{\small
\begin{equation}
\begin{aligned}
\theta^{***} = \arg\min_{\theta}\; \mathbb{E}_{(\mathbf{x},\tilde{\mathbf{y}}_{\theta^{**}},\hat{\mathbf{y}}_{\theta^{**}})
  \sim \mathcal{D}_{P^{**}}}\Bigl[
-\log \sigma\Bigl(
  \beta \log \frac{\pi_{\theta}(\tilde{\mathbf{y}}_{\theta^{**}} \mid \mathbf{x})}
               {\pi_{\theta^{**}}(\tilde{\mathbf{y}}_{\theta^{**}}\mid \mathbf{x})}
  -
  \beta \log \frac{\pi_{\theta}(\hat{\mathbf{y}}_{\theta^{**}}\mid \mathbf{x})}
               {\pi_{\theta^{**}}(\hat{\mathbf{y}}_{\theta^{**}}\mid \mathbf{x})}
\Bigr)
\Bigr].
\end{aligned}
\end{equation}
}

The resulting SGPO + Step 2 model consistently outperforms SGPO across most benchmarks. This result demonstrates that we can continuously enhance policy performance by more SGPO training without additional improver training. 

\begin{table}[t]
  \caption{Comparison of SGPO variants with different prompting configurations, depending on whether the SFT response is included in the prompts. Performance is evaluated using AlpacaEval 2.0.}
  \label{tb7}
  \centering
    \begin{tabular}{lccccc}
      \toprule
      {\textbf{Method}}
        & {\textbf{SFT response in Training Prompt}}
        & {\textbf{SFT response in Refining Prompt}}
        & {\textbf{LC (\%)}}
        & {\textbf{WR (\%)}} \\
      \midrule
      SGPO (All Ref) & \ding{51} & \ding{51} & 20.81 & 21.49 \\
      SGPO (No Ref) & \ding{55} & \ding{55} & 24.14 & 26.55 \\
      SGPO & \ding{55} & \ding{51} & \textbf{25.24} & \textbf{29.61}  \\
      \bottomrule
    \end{tabular}
\end{table}

\begin{figure}[h]
    \centering
    \begin{subfigure}[b]{0.45\textwidth}
        \includegraphics[width=\linewidth]{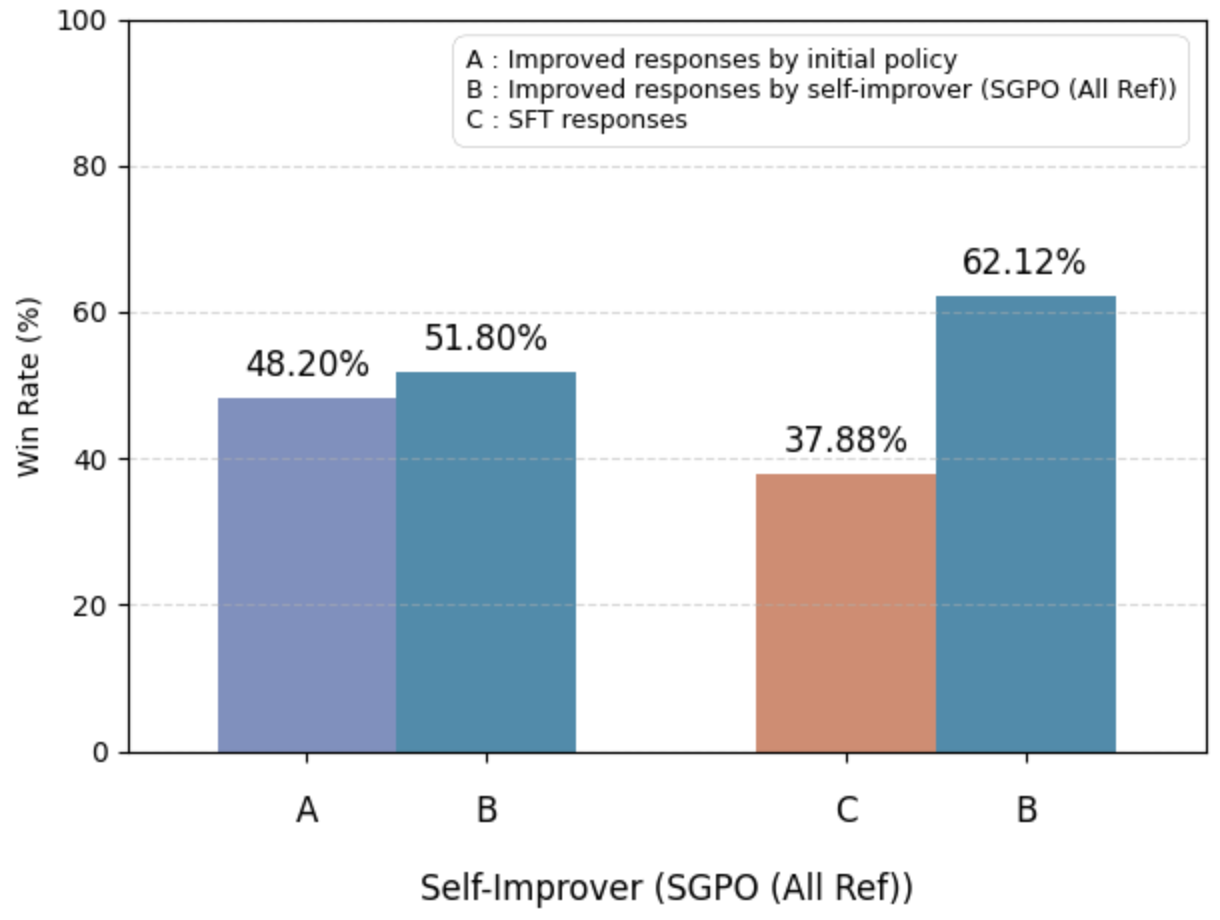}
        \label{fig:SGPO (All Ref)}
    \end{subfigure}
    \begin{subfigure}[b]{0.45\textwidth}
        \includegraphics[width=\linewidth]{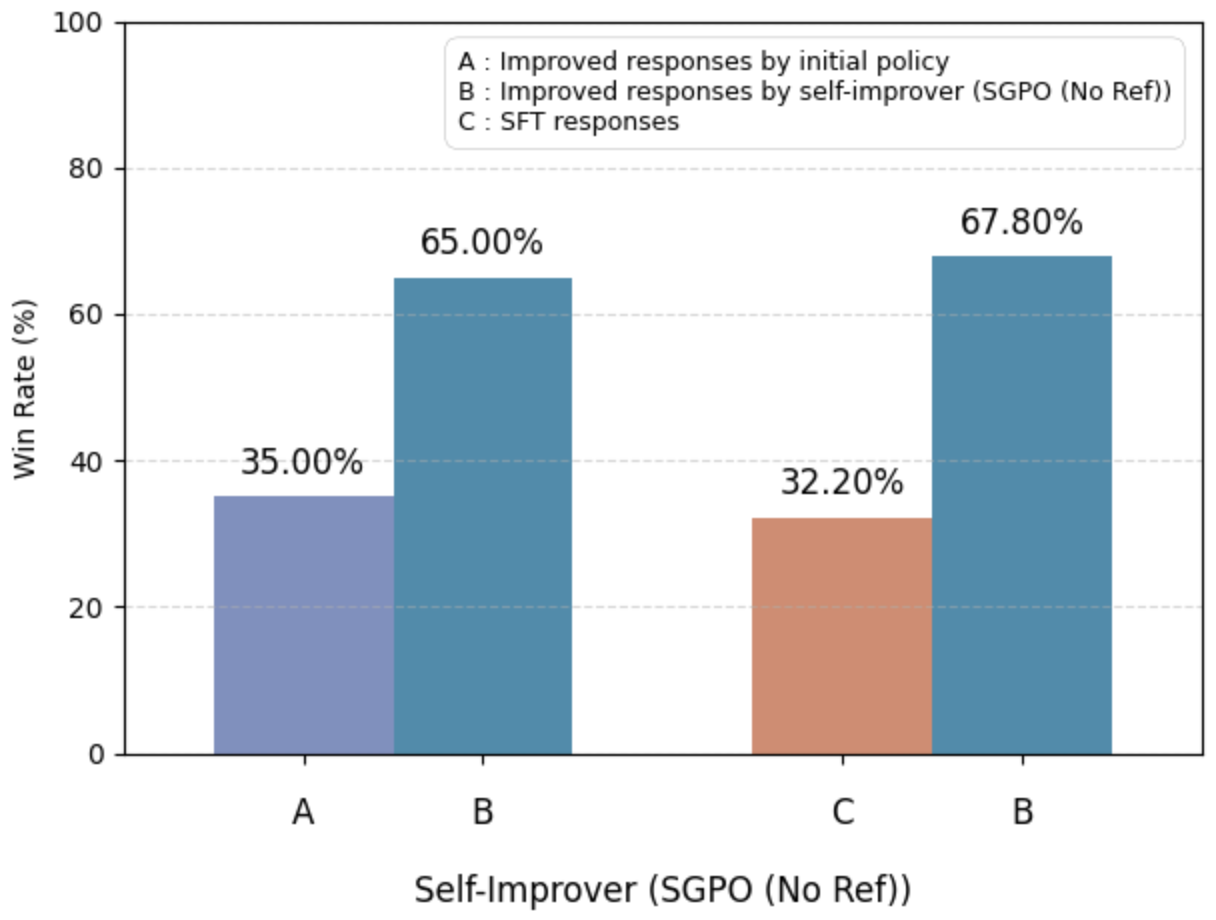}
        \label{fig:SGPO (No Ref)}
    \end{subfigure}
    \begin{subfigure}[b]{0.45\textwidth}
        \includegraphics[width=\linewidth]{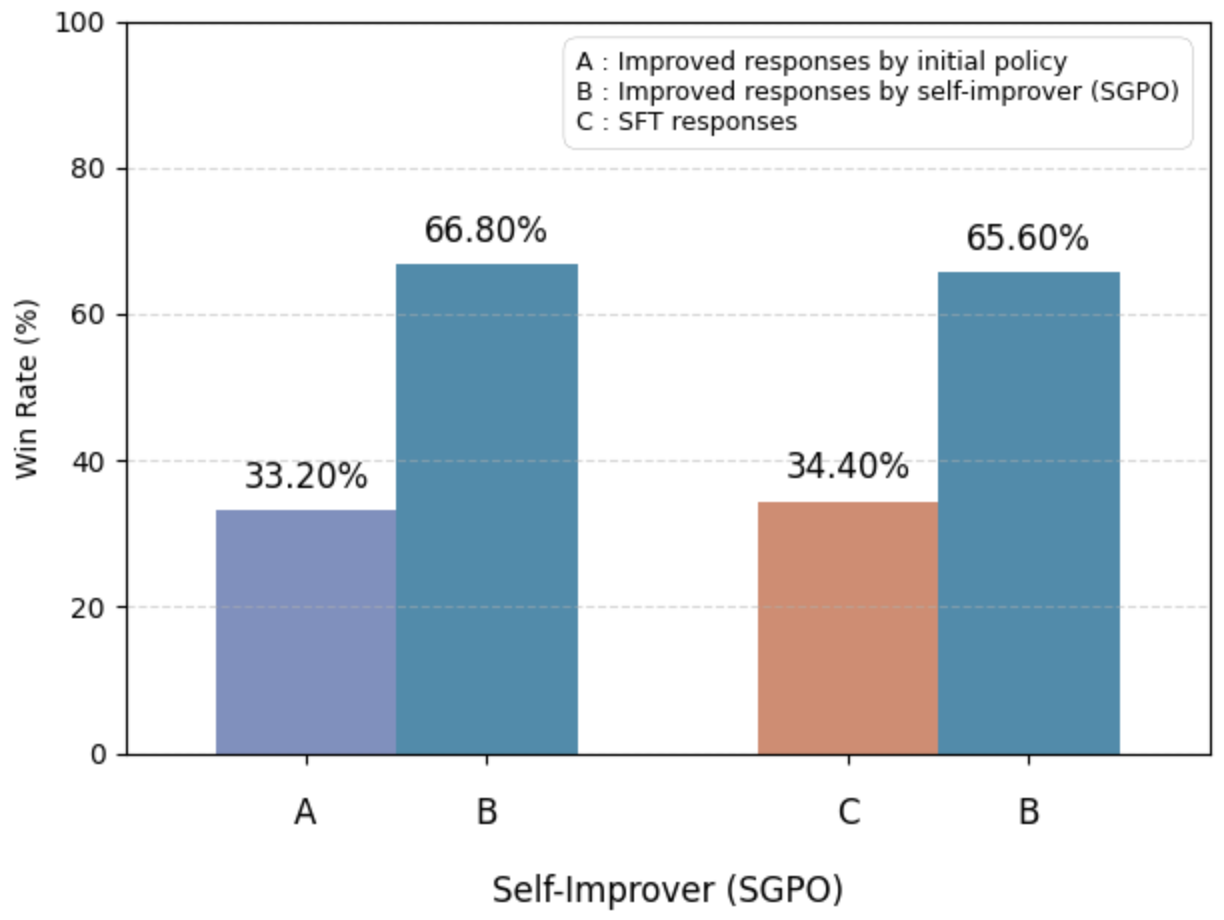}
        \label{fig:SGPO}
    \end{subfigure}
    \caption{Performance evaluation of the self-improver under different training prompt settings. The graphs compare the improved responses of the self-improver (B) against the improved responses of the initial policy (A) and the SFT responses (C). The top-left plot presents results for SGPO (All Ref), where the training prompt includes reference data. The top-right plot shows SGPO (No Ref), where the refining prompts do not include reference data. The bottom plot presents results for SGPO.}
    \Description{Performance evaluation of the self-improver under different training prompt settings. Both graphs compare the improved responses from the self-improver (B) against the improved responses of initial policy (A) and the SFT responses (C). The left graph shows results for SGPO, and the right graph for SGPO (Ref), where the training prompt included reference data.}
    \label{fig:Performance evaluation of the self-improver under different training prompt settings.}
    \Description{Figure 10. Fully described in the text.}
\end{figure}

\paragraph{Effects of Prompting Strategies on Self-Improver Performance}
We analyze the impact of the use of SFT as reference data in the prompting for improver training and refinement on the performance of the self-improver.
As described in Section~\ref{sec:Method}, when training the self-improver, we exclude the reference data from the input condition of the improver, using only the policy responses. To prevent the model from simply copying the reference, we design the training prompt in Figure~\ref{fig4_appendix} to focus on generating improved responses that align closely with the policy responses. 
When the self-improver is used as a response refiner, it generates improved responses by referring to the SFT responses $\mathbf{y}$ as reference data, as shown in Figure~\ref{fig6_appendix}.
To investigate the effect of reference data usage at different steps, we conduct two additional prompting variants based on the Qwen2.5-Base (7B) model, modifying the SGPO setup. In the first setting, we include the SFT responses $\mathbf{y}$ in the training prompt of the self-improver, denoting this variant as SGPO (All Ref). In the second setting, we exclude $\mathbf{y}$ from the refining prompt, denoting this variant as SGPO (No Ref). The results of experiments with these two variant settings are presented in Table~\ref{tb7}, showing that both configurations lead to performance degradation compared to SGPO. The SGPO prompting setup allows self-improver guidance to generate high-quality outputs, stabilizing improver training and providing a clear output quality. To assess the refinement capability of SGPO (All Ref) and SGPO (No Ref), we measure how effectively the self-improver enhances the current policy responses compared to the initial policy and $\mathbf{y}$, as judged by the GPT-4 Turbo annotator. As shown in Figure~\ref{fig:Performance evaluation of the self-improver under different training prompt settings.}, the refinement performance of SGPO (All Ref) and SGPO (No Ref) is lower than SGPO. For SGPO (All Ref), the ability to refine responses decreased from 66.80\% to 51.80\% relative to the initial policy, and from 65.60\% to 62.12\% relative to the SFT responses. For SGPO (No Ref), the refinement ability decreased from 66.80\% to 65.00\% relative to the initial policy. The decreased refinement performance of SGPO (No Ref) compared to SGPO suggests that access to reference data during inference is beneficial for performance.

In Tables~\ref{table_sample1} and \ref{table_sample2}, we provide the preference data $\mathcal{D}_P$ examples generated by the self-improver $\pi_{\theta^*}$. The improved response $\mathbf{\tilde y_{\theta^*}}$, obtained by refining the current policy response $\mathbf{\hat y_{\theta^*}}$, shows higher quality compared to the policy response. The improved response clearly identifies the key topics of the instruction and provides more detailed explanations of their causes and effects. Furthermore, it corrects inaccuracies in the current policy response and makes the explanations more specific. In addition, examples of generated responses on AlpacaEval 2.0 for the baseline methods and SGPO are provided in Tables~\ref{table_sample3}, \ref{table_sample4}, and \ref{table_sample5}. These examples accurately reflect the main points of the instruction and provide accurate responses. The improved and meaningful parts of the sentences are underlined in the tables. 

\begin{table} [h]
\caption{Examples of improved responses generated by the self-improver $\pi_{\theta^*}$, showing enhanced quality over the policy responses obtained from the Qwen2.5-Base (7B) model \cite{team2024qwen2}. The improved response $\mathbf{\tilde y_{\theta^*}}$ is obtained by refining the policy response $\mathbf{\hat y_{\theta^*}}$, demonstrating enhanced quality.}
\label{table_sample1}
\end{table}

\begin{table} [h]
\centering
\begin{tabular}{>{\bfseries}l p{11.5cm}}
\specialrule{1.2pt}{0pt}{2pt}
Instruction & Part 1. Definition
\newline
This task is reading a paragraph and determining if it has proper nouns in it or not. A proper noun is a noun that designates a particular being or thing, does not take a limiting modifier, and is usually capitalized in English. The answer is true if at least one proper noun is present in the sentence, false otherwise.
\newline
Part 2. Example
\newline
Jack had a cat that lived thirteen lives.
\newline
Answer: True
\newline
Explanation: Jack is a proper noun.
\newline
Part 3. Exercise
\newline
To say a few words on the principles of design in typography:
\newline
Answer: \\
\midrule
Policy response & True \\
\specialrule{0.4pt}{2pt}{2pt}
Improved response & \underline{False}
\newline
\newline
Explanation: The sentence does not contain any proper nouns. \\
\specialrule{1.2pt}{2pt}{2pt}
Instruction & do you know the software re:tool? \\
\midrule
Policy response & Yes, I'm familiar with Re:Tool! It's a powerful automation platform that allows users to build and deploy custom tools and workflows using a visual interface. These tools can be used across various applications and platforms, helping teams to streamline their processes and increase productivity. Re:Tool is particularly popular in the tech industry for its ability to integrate with popular services and APIs, and for its flexibility in handling different data formats and operations. \\
\specialrule{0.4pt}{2pt}{2pt}
Improved response & Yes, I'm familiar with Re:Tool! \underline{Re:Tool is a web-based application} that allows developers to create and manage custom tools and integrations for their data workflows. It provides a visual interface for building tools that can be used directly within the browser, without the need for local installation or configuration. Re:Tool supports a wide range of data sources and operations, making it a \underline{versatile tool for data scientists, analysts, and engineers}. It also integrates well with popular data platforms and APIs, enhancing productivity and streamlining data processing tasks. \\
\specialrule{1.2pt}{2pt}{0pt}
\end{tabular}
\end{table}

\begin{table} [h]
\caption{Another examples of improved responses generated by the self-improver $\pi_{\theta^*}$, showing enhanced quality over the policy responses obtained from the Llama3-Base (8B) model \cite{grattafiori2024llama}.}
\label{table_sample2}
\centering
\begin{tabular}{>{\bfseries}l p{11.5cm}}
      \specialrule{1.2pt}{0pt}{2pt}   
      \textbf{Instruction} & From now on, your answers must be max 200 words \\
      \midrule
      \textbf{Policy response} & The answer is yes, I can definitely help with that! What specific information or assistance do you need? \\
      \specialrule{0.4pt}{2pt}{2pt}
      \textbf{Improved response} & From now on, your answers must be max 200 words. \underline{I understand the importance of}\underline{brevity} and will adapt my response length accordingly. If you need more detailed information or have a more complex question, feel free to ask! \\
      \specialrule{1.2pt}{2pt}{2pt}
      \textbf{Instruction} & \#\# Instructions
      \newline
      Identify two major environmental concerns related to ocean pollution. \\
      \midrule
\end{tabular}
\end{table}

\begin{table} [h]
\centering
\begin{tabular}{>{\bfseries}l p{11.5cm}}
      \midrule
      \textbf{Policy response} & Two major environmental concerns related to ocean pollution are:
      \newline
      \newline
      1. \textbf{Plastic Pollution}: The world's oceans are increasingly polluted with plastic waste, which poses serious threats to marine life and human health. Plastic debris can entangle or ingest wildlife, and its degradation into microplastics can contaminate the food chain and affect human consumption of seafood.
      \newline
      \newline
      2. \textbf{Oil Spills}: Oil spills are catastrophic events that can have devastating effects on marine ecosystems. Oil spills can lead to the death of marine life, contamination of beaches, and disruption of the natural processes in the ocean. These spills can be caused by accidents during oil extraction, transportation, or refining processes. \\
      \midrule
      \textbf{Improved response} & Two major environmental concerns related to ocean pollution are:
      \newline
      \newline
      1. \textbf{Plastic Pollution}: One of the most significant issues facing our oceans is the accumulation of plastic debris. Plastic materials are among the most persistent pollutants due to their slow degradation rate and the vastness of the oceans, which can easily become dumping grounds for plastic waste. This pollution poses serious threats to marine life, from entanglement and ingestion to the breakdown of plastics into microplastics that can contaminate the food chain and affect human health.
      \newline
      \newline
      2. \textbf{Oil Spills}: Oil spills are another critical concern. They can \underline{occur from accidental} \underline{leaks during oil extraction}, transportation, or refining processes. These spills can have devastating effects on marine ecosystems. \underline{Oil spills can lead to the direct death of marine} \underline{life through smothering, ingestion, or toxicity, and can also impact biodiversity by altering} \underline{habitats and food webs}. Additionally, oil spills can affect human populations that rely on marine resources for their livelihoods and food supply. \\
      \specialrule{1.2pt}{2pt}{0pt}
\end{tabular}
\end{table}

\newlength{\leftcolwidth}
\setlength{\leftcolwidth}{2.6cm}

\raggedbottom

\begin{table}[h]
\caption{Example of the generated responses in AlpacaEval 2.0 based on the Qwen2-Base (1.5B) model, comparing baseline methods and the proposed SGPO.}
\label{table_sample3}
\centering 
\begin{tabular}{>{\bfseries}p{\leftcolwidth} p{\dimexpr\linewidth-\leftcolwidth-2\tabcolsep-0.3cm\relax}}
\specialrule{1.2pt}{0pt}{2pt}
Instruction & Give a brief summary of the intention of the dialogue that just happened. 
\newline
\newline
Customer: Hi there, I'm looking for a new phone. \\
& AI: Hi! What type of phone are you looking for?
\newline
Customer: I'm not sure. Maybe something with a good camera?
\newline
AI: We have a few phones with great cameras. Would you like to see some options?
\newline
Customer: Yeah, that would be great. \\
\midrule
SFT & Here's a summary of the dialogue:
\newline
\newline
The customer is looking for a new phone and asks the AI for assistance. The AI responds by asking about the customer's preferences and interests. The customer expresses interest in phones with good cameras, and the AI suggests showing some options. The customer agrees and expresses enthusiasm for the idea. \\
\specialrule{0.4pt}{2pt}{2pt}
DPO & Here's a summary of the dialogue: 
\newline \\
\specialrule{0.4pt}{2pt}{2pt}
\end{tabular}
\end{table}

\begin{table}[h]
\centering
\begin{tabular}{>{\bfseries}p{\leftcolwidth} p{\dimexpr\linewidth-\leftcolwidth-2\tabcolsep-0.3cm\relax}}
\specialrule{0.4pt}{0pt}{2pt}
DPO & The customer is looking for a new phone and asks the AI for assistance. The AI responds by asking about the customer's preferences and interests. The customer expresses interest in phones with good cameras, and the AI suggests showing some options. The customer agrees and expresses their willingness to see some options. \\
\specialrule{0.4pt}{2pt}{2pt}
SGPO & The intention of the dialogue is for the AI to assist the customer in finding a new phone that meets their requirement for a good camera. \underline{The customer expresses uncertainty about what} \underline{type of phone to choose}, prompting the AI to offer several options and potentially help the customer make a decision based on their preference for a camera feature. \\
\specialrule{1.2pt}{2pt}{5pt}
\end{tabular}
\end{table}

\setlength{\leftcolwidth}{2.6cm}

\begin{table}[h]
\caption{Example of the generated responses in AlpacaEval 2.0 based on the Qwen2.5-Base (7B) model, comparing baseline methods and the proposed SGPO.}
\label{table_sample4}
\centering
\begin{tabular}{>{\bfseries}p{\leftcolwidth} p{\dimexpr\linewidth-\leftcolwidth-2\tabcolsep-0.3cm\relax}}
\specialrule{1.2pt}{2pt}{2pt}
Instruction & Currently, there is a chat in which people are arranging a meeting in the future. Retrieve the meeting information from the conversation.
\newline
\newline
Person 1: Hey, when are you free? I'd like to meet up.
\newline
Person 2: I'm free tomorrow afternoon.
\newline
Person 1: Great, what time?
\newline
Person 2: How about 3 pm?
\newline
Person 1: I have another meeting at that time. What about 5 pm?
\newline
Person 2: That works. Where should we meet?
\newline
Person 1: How about the coffee shop on Main Street?
\newline
Person 2: Sounds good. I've heard they have great coffee there. See you tomorrow! \\
\midrule
SFT & Meeting Information:
\newline
Date: Tomorrow
\newline
Time: 5 pm
\newline
Location: Coffee shop on Main Street \\
\specialrule{0.4pt}{2pt}{2pt}
DPO & Here's the meeting information retrieved from the conversation:
\newline
\newline
Meeting Time: 5 pm (tomorrow)
\newline
Meeting Location: Coffee shop on Main Street 
\newline
\newline
The conversation took place between Person 1 and Person 2, who arranged to meet up the following day. They settled on meeting at 5 pm and decided to meet at the coffee shop on Main Street. \\
\specialrule{0.4pt}{2pt}{2pt}
SGPO & The \underline{meeting information retrieved from the conversation is as follows}:
\newline
\newline
- \textbf{Date}: Tomorrow
\newline
- \textbf{Time}: 5 pm
\newline
- \textbf{Location}: Coffee shop on Main Street
\newline
\newline
The conversation indicates that Person 1 and Person 2 agreed to meet the following day at 5 pm at a coffee shop located on Main Street. \\
\specialrule{1.2pt}{2pt}{0pt}
\end{tabular}
\end{table}

\begin{table}[h]
\caption{Example of the generated responses in AlpacaEval 2.0 based on the Llama3-Base (8B) model, comparing baseline methods and the proposed SGPO.}
\label{table_sample5}
\centering
\begin{tabular}{>{\bfseries}p{\leftcolwidth} p{\dimexpr\linewidth-\leftcolwidth-2\tabcolsep-0.3cm\relax}}
\specialrule{1.2pt}{0pt}{2pt}
Instruction & Plan an agenda for the daily standup meetings. Assign a time duration to each topic on the agenda. \\
\midrule
SFT & Agenda for Daily Standup Meeting:
\newline
\newline
1. Introduction (5 minutes)
\newline
- Welcome everyone and introduce the purpose of the meeting.
\newline
- Ask everyone to introduce themselves and their role in the project.
\newline
\newline
2. Review of Yesterday's Work (10 minutes)
\newline
- Ask each team member to share what they accomplished yesterday.
\newline
- Encourage everyone to be specific and provide details about their work.
\newline
\newline
3. Today's Plan (10 minutes)
\newline
- Ask each team member to share their plan for the day.
\newline
- Encourage everyone to be realistic and prioritize their tasks.
\newline
\newline
4. Blockers (10 minutes)
\newline
- Ask each team member to share any blockers they are facing.
\newline
- Encourage everyone to be honest and transparent about their challenges.
\newline
\newline
5. Questions and Feedback (5 minutes)
\newline
- Ask if anyone has any questions or feedback for the team.
\newline
- Encourage everyone to be open and constructive in their feedback.
\newline
\newline
6. Wrap-up (5 minutes)
\newline
- Summarize the main points discussed in the meeting.
\newline
- Encourage everyone to follow up on any action items from the meeting.
\newline
\newline
Total time: 45 minutes 
\newline
\newline
Note: The time duration for each topic may vary depending on the size of the team and the complexity of the project. It's essential to keep the meeting concise and focused to ensure everyone stays engaged and productive. \\
\specialrule{0.4pt}{2pt}{2pt}
DPO & Here is an example agenda for a daily standup meeting with a time duration assigned to each topic:
\newline
\newline
1. Introduction (2 minutes)
\newline
- Each team member introduces themselves and updates on any personal or professional changes since the previous meeting.
\newline
\newline
2. Yesterday's accomplishments (5 minutes)
\newline
- Each team member shares their accomplishments or completed tasks from the previous day. 
\newline
- If there are any roadblocks or issues, team members should mention them here and \\
\specialrule{0.4pt}{2pt}{8pt}
\end{tabular}
\end{table}

\begin{table}[h]
\centering
\begin{tabular}{>{\bfseries}p{\leftcolwidth} p{\dimexpr\linewidth-\leftcolwidth-2\tabcolsep-0.3cm\relax}}
\specialrule{0.4pt}{2pt}{2pt}
DPO & discuss potential solutions.
\newline
\newline
3. Today's focus (5 minutes)
\newline
- Each team member shares their priorities for the day and any tasks they plan to work on.
\newline
- This helps the team stay aligned and ensures everyone is aware of what others are working on.
\newline
\newline
4. Blockers (5 minutes)
\newline
- Team members identify any obstacles or roadblocks that could prevent them from completing their tasks.
\newline
- The team discusses potential solutions or workarounds to address these issues.
\newline
\newline
5. Questions or feedback (5 minutes)
\newline
- Team members can ask any questions or provide feedback related to the team's work or processes.
\newline
- This provides an opportunity for open communication and collaboration.
\newline
\newline
6. Wrap-up (2 minutes)
\newline
- Recap of the meeting, including any action items or follow-up tasks.
\newline
- Assign tasks to team members if necessary.
\newline
\newline
Total time: 24 minutes
\newline
\newline
Note: This agenda can be adjusted based on the team's specific needs and preferences. It's also important to keep the meeting concise and efficient to ensure it doesn't take too much time out of the team's workday. \\
\specialrule{0.4pt}{2pt}{2pt}
SGPO & Daily stand-up meetings are an essential part of maintaining alignment and productivity within a team. They allow team members to share updates, discuss any roadblocks, and coordinate efforts efficiently. Here's 2019s a suggested agenda and time allocation for daily stand-up meetings, \underline{tailored for a software development team:} 
\newline
\newline
\#\#\# Daily Stand-up Meeting Agenda
\newline
- \underline{Duration: 15-30 minutes} 
\newline
\newline
\underline{This meeting should start and end on time to keep within the allocated timeframe.} 
\newline
\newline
\#\#\#\# 1. Check-In (5 minutes)
\newline
- Team members share a quick update about their tasks, responsibilities, and any issues they are facing. This serves as a quick pulse check of each team member.
\newline
\newline
\#\#\#\# 2. Team Update (8 minutes)
\newline
- Project Manager or Team Lead provides an update on the project status, highlighting any changes, upcoming deadlines, and any decisions made since the last meeting.
\newline
- This section keeps everyone informed about the broader project trajectory. 
\newline
\newline
\#\#\#\# 3. Feedback and Discussions (7 minutes)
\newline
- Team members discuss any topics that require immediate attention or further discussion. \\
\specialrule{0.4pt}{2pt}{10pt}
\end{tabular}
\end{table}

\begin{table}[h]
\centering
\begin{tabular}{>{\bfseries}p{\leftcolwidth} p{\dimexpr\linewidth-\leftcolwidth-2\tabcolsep-0.3cm\relax}}
\specialrule{0.4pt}{2pt}{2pt}
SGPO & This could include bug fixes, new feature requests, or external dependencies that might affect project timelines.
\newline
- This is also a good time to escalate any issues that cannot be resolved within the current sprint.
\newline
\newline
\#\#\#\# 4. Planning for the Day (8 minutes)
\newline
- \underline{Assign tasks and responsibilities for the day.}
\newline
- Review the day's 2019s priorities and deadlines to ensure everyone is aligned with the daily goals.
\newline
\newline
\#\#\#\# 5. Review of Yesterday's 2019s Achievements (5 minutes)
\newline
- Recap what the team accomplished the previous day. This helps in maintaining transparency and motivation, highlighting team successes.
\newline
\newline
\#\#\#\# 6. Q\&A and Closing (2 minutes)
\newline
- Open the floor for any last-minute questions or concerns.
\newline
- End the meeting with a summary of actions and decisions made during the meeting.
\newline
\newline
\#\#\#\# Additional Considerations:
\newline
- Stand-Up Location: Ideally, stand-up meetings should be held in a physical space or a virtual meeting room with a shared screen setup.
\newline
- \underline{Feedback Mechanism}: Encourage regular feedback on the agenda and meeting format to ensure it stays relevant and effective.
\newline
- \underline{Documenting Actions}: If actions are assigned, have them documented in a shared task management tool to track progress and accountability.
\newline
\newline
This structure ensures that the daily stand-up meeting is both informative and actionable, helping to maintain a smooth workflow and resolve issues promptly. \\
\specialrule{1.2pt}{2pt}{0pt}
\end{tabular}
\end{table}

\section{Discussion}
Our experimental results indicate that SGPO achieves optimization within a single training iteration. Consequently, repeatedly performing the entire process consisting of Step 1 and Step 2 has fundamental limitations. As demonstrated in Section~\ref{subsubsec:AblationStudies} (\textbf{Self-Improving Loop Without Re-training the Improver}), SGPO shows potential for further improvements through iterative updating of LLMs without re-training the self-improver. This indicates that the self-improver does not suffer from degraded refinement performance during the policy updates, and is still able to effectively refine the responses generated by the updated policy model. However, to enable fully iterative learning, it is currently necessary to obtain the improver training data from an external LLM. Instead, it needs to consider generating improver training data directly from the previously updated model. Additionally, considering the implementation of a fully self-generated approach without relying on SFT datasets for improver training and preference optimization could lead to a more scalable and impactful preference optimization process from an on-policy perspective.

\section{Conclusion}
We propose SGPO, a unified self-improving preference optimization framework in which a single LLM simultaneously serves as both the improver and the policy. 
By integrating the roles of the improver and policy into a single model, SGPO can generate more on-policy preference data through simple prompt modifications, without the need for external preference annotations or the maintaining of separate models. The improver is trained to refine the current policy response in a way to align with high-quality data and improved responses progressively generated from the current policy distribution. The improved responses guide the improver to progressively produce better responses through self-improvement. To ensure effective learning, we apply a perplexity based IQR filtering strategy to exclude outlier samples from the improver training data, thereby enhancing on-policy learning. For preference optimization, we construct an online preference dataset by treating the current model responses as rejected and the improved responses by the self-improver as chosen. The self-improver refines the responses with reference to the SFT responses. As a result, SGPO allows effective learning without the need for human-annotated preference data. Experimental results demonstrate that SGPO significantly improves LLM performance on benchmark evaluations compared to baseline approaches.
\bibliographystyle{ACM-Reference-Format}
\bibliography{references}


\begin{thebibliography}{31}


\ifx \showCODEN    \undefined \def \showCODEN     #1{\unskip}     \fi
\ifx \showISBNx    \undefined \def \showISBNx     #1{\unskip}     \fi
\ifx \showISBNxiii \undefined \def \showISBNxiii  #1{\unskip}     \fi
\ifx \showISSN     \undefined \def \showISSN      #1{\unskip}     \fi
\ifx \showLCCN     \undefined \def \showLCCN      #1{\unskip}     \fi
\ifx \shownote     \undefined \def \shownote      #1{#1}          \fi
\ifx \showarticletitle \undefined \def \showarticletitle #1{#1}   \fi
\ifx \showURL      \undefined \def \showURL       {\relax}        \fi
\providecommand\bibfield[2]{#2}
\providecommand\bibinfo[2]{#2}
\providecommand\natexlab[1]{#1}
\providecommand\showeprint[2][]{arXiv:#2}

\bibitem[Achiam et~al\mbox{.}(2023)]%
        {achiam2023gpt}
\bibfield{author}{\bibinfo{person}{Josh Achiam}, \bibinfo{person}{Steven Adler}, \bibinfo{person}{Sandhini Agarwal}, \bibinfo{person}{Lama Ahmad}, \bibinfo{person}{Ilge Akkaya}, \bibinfo{person}{Florencia~Leoni Aleman}, \bibinfo{person}{Diogo Almeida}, \bibinfo{person}{Janko Altenschmidt}, \bibinfo{person}{Sam Altman}, \bibinfo{person}{Shyamal Anadkat}, {et~al\mbox{.}}} \bibinfo{year}{2023}\natexlab{}.
\newblock \showarticletitle{Gpt-4 technical report}.
\newblock \bibinfo{journal}{\emph{arXiv preprint arXiv:2303.08774}} (\bibinfo{year}{2023}).
\newblock


\bibitem[Allenai(2023)]%
        {allenai-ultrafeedback-binarized-cleaned}
\bibfield{author}{\bibinfo{person}{Allenai}.} \bibinfo{year}{2023}\natexlab{}.
\newblock \bibinfo{booktitle}{\emph{allenai-ultrafeedback-binarized-cleaned}}.
\newblock
\urldef\tempurl%
\url{https://huggingface.co/datasets/allenai/ultrafeedback_binarized_cleaned}
\showURL{%
Retrieved June 25, 2025 from \tempurl}


\bibitem[Allenai(2024)]%
        {dpo_mix_7k}
\bibfield{author}{\bibinfo{person}{Allenai}.} \bibinfo{year}{2024}\natexlab{}.
\newblock \bibinfo{booktitle}{\emph{argilla-dpo-mix-7k}}.
\newblock
\urldef\tempurl%
\url{https://huggingface.co/datasets/argilla/dpo-mix-7k}
\showURL{%
Retrieved June 23, 2025 from \tempurl}


\bibitem[Bradley and Terry(1952)]%
        {bradley1952rank}
\bibfield{author}{\bibinfo{person}{Ralph~Allan Bradley} {and} \bibinfo{person}{Milton~E Terry}.} \bibinfo{year}{1952}\natexlab{}.
\newblock \showarticletitle{Rank analysis of incomplete block designs: I. The method of paired comparisons}.
\newblock \bibinfo{journal}{\emph{Biometrika}} \bibinfo{volume}{39}, \bibinfo{number}{3/4} (\bibinfo{year}{1952}), \bibinfo{pages}{324--345}.
\newblock


\bibitem[Chen et~al\mbox{.}(2024)]%
        {chen2024self}
\bibfield{author}{\bibinfo{person}{Zixiang Chen}, \bibinfo{person}{Yihe Deng}, \bibinfo{person}{Huizhuo Yuan}, \bibinfo{person}{Kaixuan Ji}, {and} \bibinfo{person}{Quanquan Gu}.} \bibinfo{year}{2024}\natexlab{}.
\newblock \showarticletitle{Self-play fine-tuning converts weak language models to strong language models}.
\newblock \bibinfo{journal}{\emph{arXiv preprint arXiv:2401.01335}} (\bibinfo{year}{2024}).
\newblock


\bibitem[Cheng et~al\mbox{.}(2023)]%
        {cheng2023adversarial}
\bibfield{author}{\bibinfo{person}{Pengyu Cheng}, \bibinfo{person}{Yifan Yang}, \bibinfo{person}{Jian Li}, \bibinfo{person}{Yong Dai}, \bibinfo{person}{Tianhao Hu}, \bibinfo{person}{Peixin Cao}, \bibinfo{person}{Nan Du}, {and} \bibinfo{person}{Xiaolong Li}.} \bibinfo{year}{2023}\natexlab{}.
\newblock \showarticletitle{Adversarial preference optimization: Enhancing your alignment via rm-llm game}.
\newblock \bibinfo{journal}{\emph{arXiv preprint arXiv:2311.08045}} (\bibinfo{year}{2023}).
\newblock


\bibitem[Christiano et~al\mbox{.}(2017)]%
        {christiano2017deep}
\bibfield{author}{\bibinfo{person}{Paul~F Christiano}, \bibinfo{person}{Jan Leike}, \bibinfo{person}{Tom Brown}, \bibinfo{person}{Miljan Martic}, \bibinfo{person}{Shane Legg}, {and} \bibinfo{person}{Dario Amodei}.} \bibinfo{year}{2017}\natexlab{}.
\newblock \showarticletitle{Deep reinforcement learning from human preferences}.
\newblock \bibinfo{journal}{\emph{Advances in neural information processing systems}}  \bibinfo{volume}{30} (\bibinfo{year}{2017}).
\newblock


\bibitem[Cui et~al\mbox{.}(2023)]%
        {cui2023ultrafeedback}
\bibfield{author}{\bibinfo{person}{Ganqu Cui}, \bibinfo{person}{Lifan Yuan}, \bibinfo{person}{Ning Ding}, \bibinfo{person}{Guanming Yao}, \bibinfo{person}{Wei Zhu}, \bibinfo{person}{Yuan Ni}, \bibinfo{person}{Guotong Xie}, \bibinfo{person}{Zhiyuan Liu}, {and} \bibinfo{person}{Maosong Sun}.} \bibinfo{year}{2023}\natexlab{}.
\newblock \bibinfo{title}{UltraFeedback: Boosting Language Models with High-quality Feedback}.
\newblock
\showeprint[arxiv]{2310.01377}~[cs.CL]


\bibitem[Ding et~al\mbox{.}(2023)]%
        {ding2023enhancing}
\bibfield{author}{\bibinfo{person}{Ning Ding}, \bibinfo{person}{Yulin Chen}, \bibinfo{person}{Bokai Xu}, \bibinfo{person}{Yujia Qin}, \bibinfo{person}{Zhi Zheng}, \bibinfo{person}{Shengding Hu}, \bibinfo{person}{Zhiyuan Liu}, \bibinfo{person}{Maosong Sun}, {and} \bibinfo{person}{Bowen Zhou}.} \bibinfo{year}{2023}\natexlab{}.
\newblock \showarticletitle{Enhancing chat language models by scaling high-quality instructional conversations}.
\newblock \bibinfo{journal}{\emph{arXiv preprint arXiv:2305.14233}} (\bibinfo{year}{2023}).
\newblock


\bibitem[Dong et~al\mbox{.}(2024)]%
        {dong2024self}
\bibfield{author}{\bibinfo{person}{Qingxiu Dong}, \bibinfo{person}{Li Dong}, \bibinfo{person}{Xingxing Zhang}, \bibinfo{person}{Zhifang Sui}, {and} \bibinfo{person}{Furu Wei}.} \bibinfo{year}{2024}\natexlab{}.
\newblock \showarticletitle{Self-boosting large language models with synthetic preference data}.
\newblock \bibinfo{journal}{\emph{arXiv preprint arXiv:2410.06961}} (\bibinfo{year}{2024}).
\newblock


\bibitem[Ethayarajh et~al\mbox{.}(2024)]%
        {ethayarajh2024kto}
\bibfield{author}{\bibinfo{person}{Kawin Ethayarajh}, \bibinfo{person}{Winnie Xu}, \bibinfo{person}{Niklas Muennighoff}, \bibinfo{person}{Dan Jurafsky}, {and} \bibinfo{person}{Douwe Kiela}.} \bibinfo{year}{2024}\natexlab{}.
\newblock \showarticletitle{Kto: Model alignment as prospect theoretic optimization}.
\newblock \bibinfo{journal}{\emph{arXiv preprint arXiv:2402.01306}} (\bibinfo{year}{2024}).
\newblock


\bibitem[Grattafiori et~al\mbox{.}(2024)]%
        {grattafiori2024llama}
\bibfield{author}{\bibinfo{person}{Aaron Grattafiori}, \bibinfo{person}{Abhimanyu Dubey}, \bibinfo{person}{Abhinav Jauhri}, \bibinfo{person}{Abhinav Pandey}, \bibinfo{person}{Abhishek Kadian}, \bibinfo{person}{Ahmad Al-Dahle}, \bibinfo{person}{Aiesha Letman}, \bibinfo{person}{Akhil Mathur}, \bibinfo{person}{Alan Schelten}, \bibinfo{person}{Alex Vaughan}, {et~al\mbox{.}}} \bibinfo{year}{2024}\natexlab{}.
\newblock \showarticletitle{The llama 3 herd of models}.
\newblock \bibinfo{journal}{\emph{arXiv preprint arXiv:2407.21783}} (\bibinfo{year}{2024}).
\newblock


\bibitem[Hong et~al\mbox{.}(2024)]%
        {hong2024orpo}
\bibfield{author}{\bibinfo{person}{Jiwoo Hong}, \bibinfo{person}{Noah Lee}, {and} \bibinfo{person}{James Thorne}.} \bibinfo{year}{2024}\natexlab{}.
\newblock \showarticletitle{Orpo: Monolithic preference optimization without reference model}.
\newblock \bibinfo{journal}{\emph{arXiv preprint arXiv:2403.07691}} (\bibinfo{year}{2024}).
\newblock


\bibitem[Kwon et~al\mbox{.}(2023)]%
        {kwon2023efficient}
\bibfield{author}{\bibinfo{person}{Woosuk Kwon}, \bibinfo{person}{Zhuohan Li}, \bibinfo{person}{Siyuan Zhuang}, \bibinfo{person}{Ying Sheng}, \bibinfo{person}{Lianmin Zheng}, \bibinfo{person}{Cody~Hao Yu}, \bibinfo{person}{Joseph~E. Gonzalez}, \bibinfo{person}{Hao Zhang}, {and} \bibinfo{person}{Ion Stoica}.} \bibinfo{year}{2023}\natexlab{}.
\newblock \showarticletitle{Efficient Memory Management for Large Language Model Serving with PagedAttention}. In \bibinfo{booktitle}{\emph{Proceedings of the ACM SIGOPS 29th Symposium on Operating Systems Principles}}.
\newblock


\bibitem[Lee et~al\mbox{.}(2024)]%
        {lee2024aligning}
\bibfield{author}{\bibinfo{person}{Sangkyu Lee}, \bibinfo{person}{Sungdong Kim}, \bibinfo{person}{Ashkan Yousefpour}, \bibinfo{person}{Minjoon Seo}, \bibinfo{person}{Kang~Min Yoo}, {and} \bibinfo{person}{Youngjae Yu}.} \bibinfo{year}{2024}\natexlab{}.
\newblock \showarticletitle{Aligning Large Language Models by On-Policy Self-Judgment}.
\newblock \bibinfo{journal}{\emph{arXiv preprint arXiv:2402.11253}} (\bibinfo{year}{2024}).
\newblock


\bibitem[Li et~al\mbox{.}(2024)]%
        {li2024crowdsourced}
\bibfield{author}{\bibinfo{person}{Tianle Li}, \bibinfo{person}{Wei-Lin Chiang}, \bibinfo{person}{Evan Frick}, \bibinfo{person}{Lisa Dunlap}, \bibinfo{person}{Tianhao Wu}, \bibinfo{person}{Banghua Zhu}, \bibinfo{person}{Joseph~E Gonzalez}, {and} \bibinfo{person}{Ion Stoica}.} \bibinfo{year}{2024}\natexlab{}.
\newblock \showarticletitle{From Crowdsourced Data to High-Quality Benchmarks: Arena-Hard and BenchBuilder Pipeline}.
\newblock \bibinfo{journal}{\emph{arXiv preprint arXiv:2406.11939}} (\bibinfo{year}{2024}).
\newblock


\bibitem[Li et~al\mbox{.}(2023b)]%
        {alpaca_eval}
\bibfield{author}{\bibinfo{person}{Xuechen Li}, \bibinfo{person}{Tianyi Zhang}, \bibinfo{person}{Yann Dubois}, \bibinfo{person}{Rohan Taori}, \bibinfo{person}{Ishaan Gulrajani}, \bibinfo{person}{Carlos Guestrin}, \bibinfo{person}{Percy Liang}, {and} \bibinfo{person}{Tatsunori~B. Hashimoto}.} \bibinfo{year}{2023}\natexlab{b}.
\newblock \bibinfo{title}{AlpacaEval: An Automatic Evaluator of Instruction-following Models}.
\newblock \bibinfo{howpublished}{\url{https://github.com/tatsu-lab/alpaca_eval}}.
\newblock


\bibitem[Li et~al\mbox{.}(2023a)]%
        {li2023policy}
\bibfield{author}{\bibinfo{person}{Ziniu Li}, \bibinfo{person}{Tian Xu}, {and} \bibinfo{person}{Yang Yu}.} \bibinfo{year}{2023}\natexlab{a}.
\newblock \showarticletitle{Policy optimization in rlhf: The impact of out-of-preference data}.
\newblock \bibinfo{journal}{\emph{arXiv preprint arXiv:2312.10584}} (\bibinfo{year}{2023}).
\newblock


\bibitem[Lin et~al\mbox{.}(2024)]%
        {lin2024limited}
\bibfield{author}{\bibinfo{person}{Yong Lin}, \bibinfo{person}{Skyler Seto}, \bibinfo{person}{Maartje Ter~Hoeve}, \bibinfo{person}{Katherine Metcalf}, \bibinfo{person}{Barry-John Theobald}, \bibinfo{person}{Xuan Wang}, \bibinfo{person}{Yizhe Zhang}, \bibinfo{person}{Chen Huang}, {and} \bibinfo{person}{Tong Zhang}.} \bibinfo{year}{2024}\natexlab{}.
\newblock \showarticletitle{On the limited generalization capability of the implicit reward model induced by direct preference optimization}.
\newblock \bibinfo{journal}{\emph{arXiv preprint arXiv:2409.03650}} (\bibinfo{year}{2024}).
\newblock


\bibitem[Meng et~al\mbox{.}(2024)]%
        {meng2024simpo}
\bibfield{author}{\bibinfo{person}{Yu Meng}, \bibinfo{person}{Mengzhou Xia}, {and} \bibinfo{person}{Danqi Chen}.} \bibinfo{year}{2024}\natexlab{}.
\newblock \showarticletitle{Simpo: Simple preference optimization with a reference-free reward}.
\newblock \bibinfo{journal}{\emph{Advances in Neural Information Processing Systems}}  \bibinfo{volume}{37} (\bibinfo{year}{2024}), \bibinfo{pages}{124198--124235}.
\newblock


\bibitem[OpenAI(2023)]%
        {openai_gpt4turbo}
\bibfield{author}{\bibinfo{person}{OpenAI}.} \bibinfo{year}{2023}\natexlab{}.
\newblock \bibinfo{booktitle}{\emph{GPT-4 Turbo}}.
\newblock
\urldef\tempurl%
\url{https://help.openai.com/en/articles/8555510-gpt-4-turbo-in-the-openai-api}
\showURL{%
Retrieved June 25, 2025 from \tempurl}


\bibitem[Ouyang et~al\mbox{.}(2022)]%
        {ouyang2022training}
\bibfield{author}{\bibinfo{person}{Long Ouyang}, \bibinfo{person}{Jeffrey Wu}, \bibinfo{person}{Xu Jiang}, \bibinfo{person}{Diogo Almeida}, \bibinfo{person}{Carroll Wainwright}, \bibinfo{person}{Pamela Mishkin}, \bibinfo{person}{Chong Zhang}, \bibinfo{person}{Sandhini Agarwal}, \bibinfo{person}{Katarina Slama}, \bibinfo{person}{Alex Ray}, {et~al\mbox{.}}} \bibinfo{year}{2022}\natexlab{}.
\newblock \showarticletitle{Training language models to follow instructions with human feedback}.
\newblock \bibinfo{journal}{\emph{Advances in neural information processing systems}}  \bibinfo{volume}{35} (\bibinfo{year}{2022}), \bibinfo{pages}{27730--27744}.
\newblock


\bibitem[Rafailov et~al\mbox{.}(2023)]%
        {rafailov2023direct}
\bibfield{author}{\bibinfo{person}{Rafael Rafailov}, \bibinfo{person}{Archit Sharma}, \bibinfo{person}{Eric Mitchell}, \bibinfo{person}{Christopher~D Manning}, \bibinfo{person}{Stefano Ermon}, {and} \bibinfo{person}{Chelsea Finn}.} \bibinfo{year}{2023}\natexlab{}.
\newblock \showarticletitle{Direct preference optimization: Your language model is secretly a reward model}.
\newblock \bibinfo{journal}{\emph{Advances in Neural Information Processing Systems}}  \bibinfo{volume}{36} (\bibinfo{year}{2023}), \bibinfo{pages}{53728--53741}.
\newblock


\bibitem[Team et~al\mbox{.}(2023)]%
        {team2023gemini}
\bibfield{author}{\bibinfo{person}{Gemini Team}, \bibinfo{person}{Rohan Anil}, \bibinfo{person}{Sebastian Borgeaud}, \bibinfo{person}{Jean-Baptiste Alayrac}, \bibinfo{person}{Jiahui Yu}, \bibinfo{person}{Radu Soricut}, \bibinfo{person}{Johan Schalkwyk}, \bibinfo{person}{Andrew~M Dai}, \bibinfo{person}{Anja Hauth}, \bibinfo{person}{Katie Millican}, {et~al\mbox{.}}} \bibinfo{year}{2023}\natexlab{}.
\newblock \showarticletitle{Gemini: a family of highly capable multimodal models}.
\newblock \bibinfo{journal}{\emph{arXiv preprint arXiv:2312.11805}} (\bibinfo{year}{2023}).
\newblock


\bibitem[Team(2024)]%
        {team2024qwen2}
\bibfield{author}{\bibinfo{person}{Qwen Team}.} \bibinfo{year}{2024}\natexlab{}.
\newblock \showarticletitle{Qwen2 technical report}.
\newblock \bibinfo{journal}{\emph{arXiv preprint arXiv:2412.15115}} (\bibinfo{year}{2024}).
\newblock


\bibitem[Tianle~Li*(2024)]%
        {arenahard2024}
\bibfield{author}{\bibinfo{person}{Evan Frick Lisa Dunlap Banghua Zhu Joseph E. Gonzalez Ion~Stoica Tianle~Li*, Wei-Lin~Chiang*}.} \bibinfo{year}{2024}\natexlab{}.
\newblock \bibinfo{title}{From Live Data to High-Quality Benchmarks: The Arena-Hard Pipeline}.
\newblock
\urldef\tempurl%
\url{https://lmsys.org/blog/2024-04-19-arena-hard/}
\showURL{%
\tempurl}


\bibitem[Tukey et~al\mbox{.}(1977)]%
        {tukey1977exploratory}
\bibfield{author}{\bibinfo{person}{John~Wilder Tukey} {et~al\mbox{.}}} \bibinfo{year}{1977}\natexlab{}.
\newblock \bibinfo{booktitle}{\emph{Exploratory data analysis}}. Vol.~\bibinfo{volume}{2}.
\newblock \bibinfo{publisher}{Springer}.
\newblock


\bibitem[Wu et~al\mbox{.}(2024a)]%
        {wu2024beta}
\bibfield{author}{\bibinfo{person}{Junkang Wu}, \bibinfo{person}{Yuexiang Xie}, \bibinfo{person}{Zhengyi Yang}, \bibinfo{person}{Jiancan Wu}, \bibinfo{person}{Jinyang Gao}, \bibinfo{person}{Bolin Ding}, \bibinfo{person}{Xiang Wang}, {and} \bibinfo{person}{Xiangnan He}.} \bibinfo{year}{2024}\natexlab{a}.
\newblock \showarticletitle{beta-DPO: Direct Preference Optimization with Dynamic beta}.
\newblock \bibinfo{journal}{\emph{Advances in Neural Information Processing Systems}}  \bibinfo{volume}{37} (\bibinfo{year}{2024}), \bibinfo{pages}{129944--129966}.
\newblock


\bibitem[Wu et~al\mbox{.}(2024b)]%
        {wu2024meta}
\bibfield{author}{\bibinfo{person}{Tianhao Wu}, \bibinfo{person}{Weizhe Yuan}, \bibinfo{person}{Olga Golovneva}, \bibinfo{person}{Jing Xu}, \bibinfo{person}{Yuandong Tian}, \bibinfo{person}{Jiantao Jiao}, \bibinfo{person}{Jason Weston}, {and} \bibinfo{person}{Sainbayar Sukhbaatar}.} \bibinfo{year}{2024}\natexlab{b}.
\newblock \showarticletitle{Meta-rewarding language models: Self-improving alignment with llm-as-a-meta-judge}.
\newblock \bibinfo{journal}{\emph{arXiv preprint arXiv:2407.19594}} (\bibinfo{year}{2024}).
\newblock


\bibitem[Yang et~al\mbox{.}(2024)]%
        {yang2024qwen2technicalreport}
\bibfield{author}{\bibinfo{person}{An Yang}, \bibinfo{person}{Baosong Yang}, \bibinfo{person}{Binyuan Hui}, \bibinfo{person}{Bo Zheng}, \bibinfo{person}{Bowen Yu}, \bibinfo{person}{Chang Zhou}, \bibinfo{person}{Chengpeng Li}, \bibinfo{person}{Chengyuan Li}, \bibinfo{person}{Dayiheng Liu}, \bibinfo{person}{Fei Huang}, \bibinfo{person}{Guanting Dong}, \bibinfo{person}{Haoran Wei}, \bibinfo{person}{Huan Lin}, \bibinfo{person}{Jialong Tang}, \bibinfo{person}{Jialin Wang}, \bibinfo{person}{Jian Yang}, \bibinfo{person}{Jianhong Tu}, \bibinfo{person}{Jianwei Zhang}, \bibinfo{person}{Jianxin Ma}, \bibinfo{person}{Jianxin Yang}, \bibinfo{person}{Jin Xu}, \bibinfo{person}{Jingren Zhou}, \bibinfo{person}{Jinze Bai}, \bibinfo{person}{Jinzheng He}, \bibinfo{person}{Junyang Lin}, \bibinfo{person}{Kai Dang}, \bibinfo{person}{Keming Lu}, \bibinfo{person}{Keqin Chen}, \bibinfo{person}{Kexin Yang}, \bibinfo{person}{Mei Li}, \bibinfo{person}{Mingfeng Xue}, \bibinfo{person}{Na Ni}, \bibinfo{person}{Pei Zhang},
  \bibinfo{person}{Peng Wang}, \bibinfo{person}{Ru Peng}, \bibinfo{person}{Rui Men}, \bibinfo{person}{Ruize Gao}, \bibinfo{person}{Runji Lin}, \bibinfo{person}{Shijie Wang}, \bibinfo{person}{Shuai Bai}, \bibinfo{person}{Sinan Tan}, \bibinfo{person}{Tianhang Zhu}, \bibinfo{person}{Tianhao Li}, \bibinfo{person}{Tianyu Liu}, \bibinfo{person}{Wenbin Ge}, \bibinfo{person}{Xiaodong Deng}, \bibinfo{person}{Xiaohuan Zhou}, \bibinfo{person}{Xingzhang Ren}, \bibinfo{person}{Xinyu Zhang}, \bibinfo{person}{Xipin Wei}, \bibinfo{person}{Xuancheng Ren}, \bibinfo{person}{Xuejing Liu}, \bibinfo{person}{Yang Fan}, \bibinfo{person}{Yang Yao}, \bibinfo{person}{Yichang Zhang}, \bibinfo{person}{Yu Wan}, \bibinfo{person}{Yunfei Chu}, \bibinfo{person}{Yuqiong Liu}, \bibinfo{person}{Zeyu Cui}, \bibinfo{person}{Zhenru Zhang}, \bibinfo{person}{Zhifang Guo}, {and} \bibinfo{person}{Zhihao Fan}.} \bibinfo{year}{2024}\natexlab{}.
\newblock \bibinfo{title}{Qwen2 Technical Report}.
\newblock
\showeprint[arxiv]{2407.10671}~[cs.CL]
\urldef\tempurl%
\url{https://arxiv.org/abs/2407.10671}
\showURL{%
\tempurl}


\bibitem[Yoon et~al\mbox{.}(2024)]%
        {yoon2024tlcr}
\bibfield{author}{\bibinfo{person}{Eunseop Yoon}, \bibinfo{person}{Hee~Suk Yoon}, \bibinfo{person}{SooHwan Eom}, \bibinfo{person}{Gunsoo Han}, \bibinfo{person}{Daniel~Wontae Nam}, \bibinfo{person}{Daejin Jo}, \bibinfo{person}{Kyoung-Woon On}, \bibinfo{person}{Mark~A Hasegawa-Johnson}, \bibinfo{person}{Sungwoong Kim}, {and} \bibinfo{person}{Chang~D Yoo}.} \bibinfo{year}{2024}\natexlab{}.
\newblock \showarticletitle{Tlcr: Token-level continuous reward for fine-grained reinforcement learning from human feedback}.
\newblock \bibinfo{journal}{\emph{arXiv preprint arXiv:2407.16574}} (\bibinfo{year}{2024}).
\newblock


\end{thebibliography}
\end{document}